\documentclass[10pt]{report}
\usepackage[utf8]{inputenc}
\usepackage{graphicx}
\graphicspath{ {images/} }
\usepackage{caption}
\usepackage{subcaption}
\usepackage{color}
\usepackage{enumerate}
\usepackage{float}
\usepackage{mathtools}
\usepackage{amssymb}
\usepackage{amsfonts}
\usepackage[width=160mm,top=25mm,bottom=25mm,bindingoffset=6mm]{geometry}
\usepackage{fancyhdr}
\usepackage[colorlinks=true,linkcolor=blue,citecolor=blue,urlcolor=blue]{hyperref}
\usepackage{setspace}
\renewcommand{\headrulewidth}{0pt}
\usepackage[style=phys, sorting=none, doi=false,url=false,isbn=false, firstinits=true, maxnames=999, date=year, hyperref=true, pageranges=false, articletitle=false, eprint=false]{biblatex}
\renewbibmacro{in:}{}
\AtEveryBibitem{\clearfield{issue}}
\AtEveryBibitem{\clearfield{note}}
\AtEveryBibitem{\clearfield{number}}
\DeclareFieldFormat{labelnumberwidth}{\mkbibbrackets{#1}}
\newcommand\norm[1]{\left\lVert#1\right\rVert}

\title{Applications of Machine Learning to Modelling and Analysing Dynamical Systems}
\author{Vedanta Thapar}
\date{\today}

\begin{document}
\pagenumbering{roman} 

\begin{titlepage}
    \begin{center}
        \vspace*{1cm}
        
        \Huge
        \textbf{Applications of Machine Learning to Modelling and Analysing Dynamical Systems}
        
        \vspace{0.5cm}
        \LARGE
        
        by\\
    
        \textbf{Vedanta Thapar}
        
        \vfill
        
        A dissertation submitted in partial fulfillment\\
        of the requirements for the degree of\\
        Bachelor of Science (Honours)\\
        Physics\\

        \vspace{1.8cm}

        \Large
        at \\St. Stephen's College\\ University of Delhi\\
        Year 2023\\
        \vspace{1.0cm}
        \begin{flushleft}
        \large
        This dissertation is guided by: \\
        \setlength{\parindent}{10ex}
        Dr. Abhinav Gupta, Associate Professor,\\
        Department of Physics, St. Stephen's College Delhi
        \end{flushleft}
        
    \end{center}
    
\end{titlepage}

\fancyhf{} 
\fancyhead[RO,R]{\thepage} 
\renewcommand{\headrulewidth}{0pt}

\begin{center}
    \Large
    \textbf{Applications of Machine Learning to Modelling and Analysing Dynamical Systems}
    
    \vspace{0.4cm}
    \large
    A Study of Hamiltonian Systems
    
    \vspace{0.4cm}
    \textbf{Vedanta Thapar}
    
    \vspace{0.9cm}
    \textbf{Abstract}
\end{center}
\textit{We explore the use of Physics Informed Neural Networks to analyse nonlinear Hamiltonian Dynamical Systems with a first integral of motion. In this work, we propose an architecture which combines existing Hamiltonian Neural Network structures into Adaptable Symplectic Recurrent Neural Networks which preserve Hamilton's equations as well as the symplectic structure of phase space while predicting dynamics for the entire parameter space. We find that this architecture significantly outperforms previously proposed neural networks when predicting Hamiltonian dynamics especially in potentials which contain multiple parameters. We demonstrate its robustness using the nonlinear Henon-Heiles potential under chaotic, quasiperiodic and periodic conditions.} \\

\textit{The second problem we tackle is whether we can use the high dimensional nonlinear capabilities of neural networks to predict the dynamics of a Hamiltonian system given only partial information of the same. Hence we attempt to take advantage of Long Short Term Memory networks to implement Takens' embedding theorem and construct a delay embedding of the system followed by mapping the topologically invariant attractor to the true form. We then layer this architecture with Adaptable Symplectic nets to allow for predictions which preserve the structure of Hamilton's equations. We show that this method works efficiently for single parameter potentials and provides accurate predictions even over long periods of time. }





\chapter*{Certificate}

{\large \textit{This is to certify that the dissertation titled \texttt{Applications of Machine Learning to Modelling and Analysing Dynamical Systems} is a bonafide work of \texttt{Vedanta Thapar}, who has done research under my guidance and supervision. This is being submitted to the Department of Physics, St. Stephen's College, University of Delhi, in partial fulfilment, in lieu of a course in \texttt{Physics
Hons. (567: BSc (Honours) Physics VI semester: Paper Code: 32227627.)}
To the best of my knowledge, the contents of the dissertation have not been previously
submitted to this or any other institution, for any other course and degree and this is his original work.}}

\vfill



\chapter*{Declaration}

{\large I certify that.

\begin{enumerate}
    \item \textit{The work contained in this undergraduate dissertation is original and has been completed by me, under the guidance of my supervisor.}

    \item \textit{The work has not been submitted to any other Institute for any degree or diploma.}

    \item  \textit{Whenever I have used materials (data, plots, theoretical analysis, and text) from other sources, I have given due credit to the original source and author by citing them in the text of the thesis and listing details in the Bibliography.}

    \item \textit{Whenever I have quoted written materials from other sources, I have given due credit to the sources.}
\end{enumerate}}

\vspace{2cm}

\noindent \textbf{Name: Vedanta Thapar} \\
\textbf{Place : St. Stephen’s College, University of Delhi} \\
\textbf{Roll Number : 20080567042}\\
\textbf{Registration Number : 20BPHY037}

\vfill

  

\chapter*{Acknowledgements}

The author is profoundly grateful to the advisor of this project, Dr. Abhinav Gupta, for his constant support and guidance over the last three years. The author would also like to thank Professor Ram Ramaswamy for his valuable advice and suggestions in this work.

\tableofcontents

\listoffigures


\doublespacing

\chapter{Introduction}\label{intro}
\pagenumbering{arabic} 
In the last few years, with the rapid growth of computing capabilities, the use of Machine Learning(ML) has taken the world by storm as a powerful tool for the analysis and prediction of large datasets. Neural Networks, a canonical ML model, have found great application in data rich fields such as medicine \cite{Patel2007Sep}, natural language processing \cite{9075398}, image recognition \cite{8320684}, among others, as well as a variety of uses in engineering and technology.  With the release of open source Large Language Models(LLMs) like ChatGPT, Bard, etc. this revolution has entered the daily lives of people around the world. Alongside these, scientific fields found ML as a tool to analyse the complex datasets for which developing closed form models proved difficult. Some examples of these include Material Science \cite{Wei2019Sep}, Biology \cite{Tarca2007Jun}, Computational Chemistry \cite{Goh2017Jun}, etc. \\

Since the 1990s there has been great interest in using Neural Networks to tackle problems in Physics \cite{DUCH199491}, specially in regimes where other computational methods have failed. However it was found that Neural Networks, although extremely powerful, struggle to take into account the physical laws underlying the data being fed. This led to the development of so-called \textbf{Physics Informed Neural Networks(PINNs)} \cite{RAISSI2019686} which integrated prior knowledge of physics within the structure of the neural network or the method of training the model. This has led to the emergence of several modified architectures which attempt to solve a variety of problems in fields such as Fluid Dynamics, Quantum Simulations, Geophysics, High-energy Physics, etc. The primary use of PINNs is in the so-called `some physics - some data' regime, i.e. alongside some knowledge of the underlying physics, there is some data available for the neural network to `fill' the missing pieces. This is in contrast to the `small data - lots of physics' regime where we essentially know the underlying equations and require only boundary/initial conditions; or the `big data-no physics' regime where a purely data driven approach might have greater effect. Several independent frameworks have been implemented in the last couple of years to attempt to discover physical laws with neural networks such as SciNet\cite{Iten_Metger_Wilming_del_Rio_Renner_2020}, AI Poincare \cite{PhysRevLett.126.180604}, AI Feynman \cite{aifeynman}, etc.  A complete review of some of these applications can be found in \cite{Karniadakis2021Jun}.\\

One of the primary `real-life' problems in physics is in the analysis of highly nonlinear, chaotic and complex dynamical systems such as climate, fluids, celestial mechanics, etc. In this work we look at \textbf{Hamiltonian Dynamical Systems} specifically those with time independent Hamiltonians where a first integral of motion(energy) can be found. These dynamical systems can display a wide variety of rich behaviour including periodic, quasiperiodic and chaotic motion. There have been attempts to analyse these with classic `deep' neural network methods including the classification of regular and chaotic motion \cite{Celletti_Gales_Rodriguez-Fernandez_Vasile_2022}, using resovoir computing to learn Hamiltonian Dynamics \cite{PhysRevE.104.024205}, etc. We analyse a class of nets referred to as \textbf{Hamiltonian Neural Networks(HNNs)} which encode the very structure of Hamilton's equation in their architecture. These were first proposed by Greynadus et al. \cite{Greydanus_Dzamba_Yosinski_2019} in 2019 and later modified by Han et al. \cite{Han_Glaz_Haile_Lai_2021} to allow for adaptability in the parameter space. Chen et al. \cite{Chen_Zhang_Arjovsky_Bottou_2020} extended the former architecture into so-called Symplectic Recurrent Neural Networks(SRNNs) which combined these Hamiltonian networks with leapfrog integrators to preserve the symplectic structure of phase space in Hamiltonian Systems. There has been signficant interest in this class of architectures in the last couple of years including forecasting of Hamiltonian Dynamics without canonical coordinates by layering dense networks with HNNs \cite{Choudhary2021Jan}, sparse symplectically integrated neural networks \cite{NEURIPS2020_439fca36}, Taylor series forms of SRNNs \cite{Tong2021Jul}, Nonseparable SRNNs \cite{Xiong2021Jan}, among several others. A review of some of the work up till 2022 can be found in \cite{Chen2022Feb}. \\

This work proposes a combination of the Adaptable Hamiltonian Neural Nets proposed by Han et al. \cite{Han_Glaz_Haile_Lai_2021} and SRNNs to construct a novel architecture: \textbf{Adaptable Symplectic Neural Networks(ASRNNs)}. A large portion of this dissertation demonstrates the robustness of the same. The second problem we attempt to tackle is using Neural Networks to predict a dynamical system given only partial information of its phase space. We take advantage of Long Short Term Memory (LSTM) \cite{LSTM} networks and their ability to embed dynamics in higher dimensional spaces to implement Takens' embedding theorem. They are used in conjunction with ASRNNs to enforce Hamiltonian dynamics as well. \\

The following Chapter introduces the various architectures involved in this project while Chapter \ref{chapter3} demonstrates the robustness of our proposed ASRNNs. Chapter \ref{chapter4} studies the problem of discovering dynamics from partial information, we conclude with an overview of salient results as well as potential future applications and improvements of the proposed models.

\chapter{Physics Informed Neural Networks} \label{chapter2}
\section{Overview of Neural Network Architectures}

\subsection{Multiperceptron Feedforward Neural Networks} \label{feedforward}

Multiperceptron Feedforward Neural Networks(FFNNs) are the most standard form of Neural networks that are capable of learning complex non-linear relationships between a system's input and output. They are most generally composed of a sequence of layers composed of densely connected nodes referred to as `neurons'. These neurons store scalar values that propagate through the layers of the network with the application of linear/non-linear functions.

\begin{figure}[ht]
    \centering
    \includegraphics[width=0.8\textwidth]{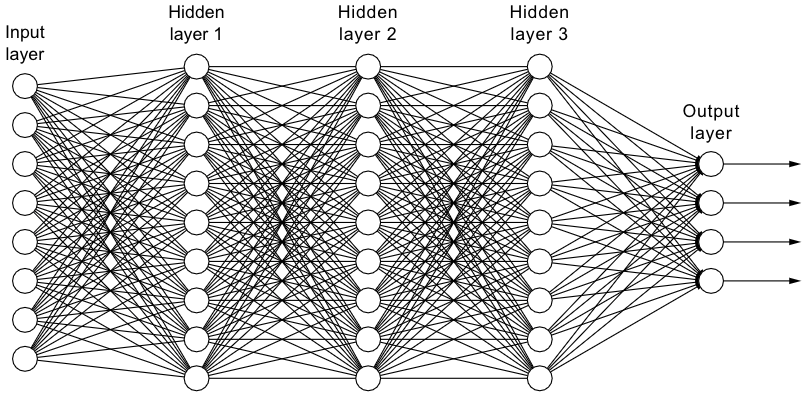}
    \caption{A Multiperceptron Feedforward Neural Network \cite{Upadhyay2021Dec}}
    \label{fig:my_label}
\end{figure}

In general, neurons within a layer do not interact with each other but are rather only connected to those in the layer immediately preceding and succeeding their own. The network is constructed as follows: an input layer composed of $n_i$ neurons takes in the required inputs of the system, followed by a sequence of `hidden layers' through which the information from the input neurons is propogated and operated upon. This sequence is connected to a final output layer of $n_{out}$ neurons which we require to be the desired outputs of the system. \\

The rule of propagation between neurons is as follows: the activation(scalar value) of the $jth$ neuron in the $(n+1)th$ layer is given by the weighted sum of all the activations of neurons in the $nth$ layer, i.e. 

\begin{equation}
    z_j^{(n+1)} = \sum_k w_{jk}^{(n+1), n}x_k^{(n)} + b_j^{(n+1)} 
\end{equation}
\begin{equation}
    x_j^{(n+1)} = f(z_j^{(n+1)}) 
\end{equation}

where $x_k^{(n)}$ represents the activation of the $kth$ neuron in the $nth$ layer, $w_{jk}^{(n+1), n}$ represent the weight associated with the connection between the $jth$ neuron of the $(n+1)th$ layer and the $kth$ neuron of the $nth$ layer. Additionally $b_j^{(n+1)}$ defines the bias of the $jth$ neuron of the $(n+1)th$ layer while $f(x)$ is a  (usually) non linear function. \\

The inputs of the system define the activations of the neurons in the input layer while the outputs define the \textbf{desired} activations of the neurons in the output layer. The task of Machine Learning is to `learn' the optimal parameters, i.e. the weights and biases that most closely map the inputs to the desired outputs. Hence the neural network essentially defines a series of non-linear operations that act on the values of the input neurons sequentially to achieve an optimum output. Networks where all neurons in sucessive layers are connected to each other, i.e. $\exists$ a non-zero weight between all neurons of successive layers, are referred to as \textbf{Dense}. \\

The task of finding the optimum parameters is transformed to a minimization problem by defining a \textit{Loss} function between the outputs of the network and desired outputs. The global minimum of this function w.r.t to all the parameters defines the optimum point. As an example consider the \textit{Mean-Squared Error} loss defined by:

\begin{equation}
    \mathcal{L}_x(\theta) = \frac{1}{N} \sum_ i\left( F_\theta(x_i) -  F(x_i) \right)^2
\end{equation}

Where $F(x_i)$ are the desired outputs for inputs $x_i$ and $F_{\theta}(x_i)$ defines the output of the neural network with parameters $\theta$. The sum is defined (in principle) over all the inputs $x_i$ with $N$ being the total number of such inputs. To form a neural network capable of learning complex non-linear mappings, the number of parameters needs to be incredibly large. finding the minimum of $\mathcal{L}_x(\theta)$ in this high dimensional parameter space becomes a computationally impossible task using traditional regression methods of vanishing derivatives. Additionally the number of training inputs to these networks is typically very large, further intensifying the task of finding this minimum. For this reason. methods of \textbf{Stochastic Gradient Descent} which involve iteratively `stepping' towards the minimum by calculating the direction of maximum descent, are employed. The idea of stochasticity is introduced to tackle the latter problem of the large number of inputs: instead of defining the cost as a sum over all inputs, we instead consider a randomly selected set of them referred to as a `batch'. The method of gradient descent requires the calculation of the gradient of the loss w.r.t to all the parameters of the neural network, a difficult task considering the large dimensionality. This problem of is partially solved with the development of the \textbf{Backpropogation} algorithm(Appendix \ref{backprop}), an efficient method to sucessively calculate the partial derivatives of the loss function w.r.t to the large number of parameters. It should be noted that the loss function is not convex, i.e. a global minimum can be difficult to find due to the existence of several local minima in its landscape. The stochasticity of the gradient descent algorithm helps us tackle this issue by allowing the algorithm to explore different regions of the parameter space, potentially escaping from poor local minima. 

\subsection{Capturing Temporal Dependencies}

FFNNs contain no method to differentiate data temporally, i.e. due to their sequential structure where all inputs are fed equally, they struggle to capture temporal or sequential dependencies in input data. This essentially means they keep no `memory' of data separated in time and are hence unable to effectively use information of `previous'(in the temporal sense) data to discover patterns. This makes them largely incapable of effectively analysing dynamical systems especially those with chaotic or time delay dynamics. This problem is solved by broadening the structure of feedforward networks to allow for feedback loops where information can be passed from one time step to the next. 

\begin{figure}
    \centering
    \includegraphics[width=\textwidth]{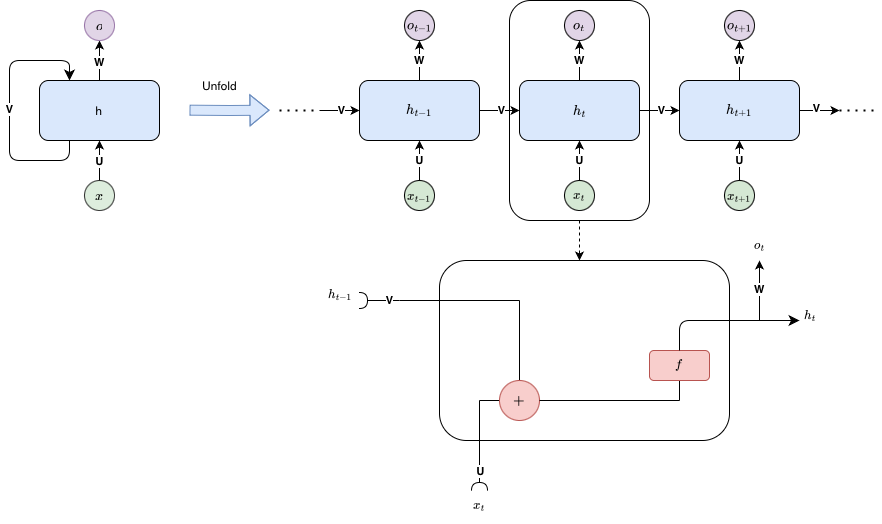}
    \caption{General Architecture of Recurrent Neural Networks}
    \label{fig:my_label}
\end{figure}

These \textbf{`Recurrent' Neural networks(RNN)} are composed of neurons with feedback loops that take in a sequence of values as input $\mathbf{x} = \{\mathbf{x}(1) \hdots \mathbf{x}(\tau) \}$. Memory is stored in an RNN cell by creating an additional, so-called `hidden' state to which inputs are mapped into. For every time step $x(t)$, the RNN unit takes in the input at that time along with the hidden state of the previous time step. The hidden state at step $t$ is defined by:

\begin{equation}
    \mathbf{h}(t) = f(\mathbf{h}(t-1), \mathbf{x}(t) ; \mathbf{\theta}) = f(\mathbf{Ux}(t) + \mathbf{Vh}(t-1) + \mathbf{b})
\end{equation}

Where $\mathbf{U}, \mathbf{V}$ are weight matrices for their respective mappings, $\mathbf{b}$ is a bias vector and $f$ is a (usually) nonlinear function as before. Common choices include the Softmax, Tanh and Rectified Linear Unit(ReLU). The output at each step $\mathbf{o}(t)$ is then

\begin{equation}
    \mathbf{o}(t) = \mathbf{Wh}(t) + \mathbf{c}
\end{equation}
\begin{equation}
    \mathbf{y}(t) = g(\mathbf{o}(t))
\end{equation}
Here $\mathbf{o}(t)$ is the output of the recurrent unit, which is usually mapped to the required output by means of a (usually) nonlinear function $g$, and $\mathbf{W}, \mathbf{c}$ are weight, bias parameters respectively. Often times, as in this work as well, the hidden state is mapped to the output by the means of a Dense neural network(instead of the function $g$) defined with the same parameters for every time step. Recurrent Neural Networks are trained by the same method as Dense ones by minimizing some $\mathcal{L}_{\theta}( F_{\theta} (\mathbf{x}), F(\mathbf{x}))$ using Gradient Descent. The Backprogpogation algorithm is extended to \textbf{Backpropogation through Time (BPTT)} through which the backward pass is not only through layers but also through time. However the biggest caveats of training RNNs is the so-called Exploding/Vanishing Gradients Problem. Due to the structure of the BPTT algorithm, which involves sucessive matrix multiplication as we go through time, if the eigenvalues of these matrices are greater/less than 1, this leads to the gradients of the loss function exploding/vanishing respectively if the series is very long. These problems can be intuitively thought of as the RNN's memory becoming oversaturated (in the case of gradient explosion) where it is unable to capture new information, or the memory is lost(in the case of gradient vanishing) and the network is unable to effectively recall information from past time steps. This problem hindered the use of RNNs initially, but was largely solved with the development of `Gated Recurrent Units(GRUs)' the most popular of which are the \textbf{Long Short Term Memory Networks(LSTM)}.

\subsubsection{Long Short Term Memory}

\begin{figure}
    \centering
    \includegraphics[width=0.6\textwidth]{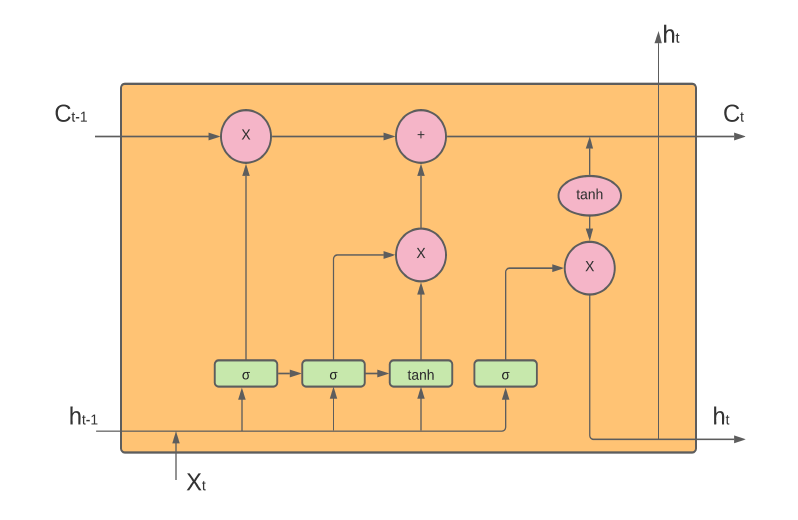}
    \caption{Architecture of General LSTM Unit}
    \label{fig:lstm}
\end{figure}

To control the flow of memory and the amount of information stored by a recurrent unit, we add additional operations(called gates) that can write/delete from the memory cell. This is often thought of as using additional neurons in addition to the general recurrent unit discussed in the previous section. These additional operations(or neurons) together with the memory cell of the RNN form a \textbf{Long Short Term Memory} unit(Fig. \ref{fig:lstm}). The LSTM takes in the same inputs as an RNN and returns the same form of output, however the internal structure is more complex.

Note that there exist several types of GRUs each defined by the additional operations they perform to control the flow of memory, an LSTM is a specific and, in general, highly robust example of the same. Most general GRUs are defined by so-called reset and update operations, an LSTM extends these with the use of $3$ gates and the inclusion of an additional `cell state' that is propagated through time much like the hidden state but with a different set of operations. This state contains an internal loop within the LSTM unit which controls how much information is propagated from the previous hidden state and the current input into the current hidden state. This is done by the $3$ gates defined as:

\begin{itemize}
    \item \textbf{Forget Gate }: This takes in the previous hidden state and current input state as input and returns a value between 0-1 that determines how much information should be retained from the previous cell state to the next one.

    \begin{equation}
        \mathbf{f}(t) = \mathbf{\sigma} \left( \mathbf{U}^f\mathbf{x}(t) + \mathbf{V}^f\mathbf{h}(t-1) + \mathbf{b}^f \right)
    \end{equation}

    Where $\mathbf{f} (t)$ is the value of the forget gate(between 0-1), $\mathbf{U}^f, \mathbf{V}^f$ are the weight matrices for the forget gate, $\sigma$ represents the sigmoid function.

    \item \textbf{Input Gate }: This acts similar to the forget gate but determines how much new information from the current input state should be added to the current cell state. It takes in the same inputs as the forget gate and returns and output between $0-1$ that determines the amount of information to be added to the cell state.

    \begin{equation}
        \mathbf{i}(t) = \mathbf{\sigma} \left( \mathbf{U}^i\mathbf{x}(t) + \mathbf{V}^i\mathbf{h}(t-1) + \mathbf{b}^i \right)
    \end{equation}

    Where the parameters are defined as in the forget gate with $\mathbf{U}^i, \mathbf{V}^i$ and $\mathbf{b}^i$ are the weight matrices and bias vectors respectively for the input gate.

    The forget and input gate together with the previous cell and hidden states determine the current cell state as follows:
    \begin{equation}
        \mathbf{c}(t) = \mathbf{f}(t) \cdot \mathbf{c}(t-1) + \mathbf{i}(t) \cdot \tanh \left( \mathbf{U}\mathbf{x}(t) + \mathbf{V}\mathbf{h}(t-1) + \mathbf{b} \right)
    \end{equation}

    Where $\mathbf{U}, \mathbf{V}$ and $\mathbf{b}$ are the weight matrices and bias vector for the LSTM cell respectively. The argument of the $\tanh$ above is often referred to as the \textit{candidate cell state} represented by $\Tilde{\mathbf{c}}(t)$.

    \item \textbf{Output Gate }: The output gate determines how much information from the current cell state is passed to the current hidden state which is eventually mapped to the output. It works much like the forget and input gates by taking the same inputs and returning a number between $0-1$ for each element of the cell state.

    \begin{equation}
        \mathbf{o}(t) = \mathbf{\sigma} \left( \mathbf{U}^o\mathbf{x}(t) + \mathbf{V}^o\mathbf{h}(t-1) + \mathbf{b}^o \right)
    \end{equation}

    The current hidden state is then defined as

    \begin{equation}
        \mathbf{h}(t) = \mathbf{o}(t) \cdot \tanh(\mathbf{c}(t))
    \end{equation}

\end{itemize}

The information flow control provided by the gates mitigates the long series problem providing the recurrent network with `Long short-term memory', i.e. the architecture has the ability to selectively remember or forget information over extended periods of time, while also retaining information from recent inputs.

\subsection{ODEnets}

Consider the problem of discovering the underlying differential equations for a dynamical system using its time series. Hence consider the dynamical system:

\begin{equation}
    \dot{\mathbf{x}} = f(\mathbf{x})
\end{equation}

Our aim with Neural Networks is to discover the function $f(\mathbf{x})$. This is done by inputting $\mathbf{x}(t)$ and requiring output $\dot{\mathbf{x}}(t)$. Network training can be performed in several ways, most simply one can feed all $\mathbf{x}$ instances at different times as separate training batches independent of each other. In this case ground truth outputs, $\dot{\mathbf{x}}(t)$, are often taken as finite differences from the time series data. In this way the trained network approximates the dynamical system as:

\begin{equation}
    \dot{\mathbf{x}} = F_{\theta}(\mathbf{x})
\end{equation}
Where $\theta$ represents the parameters of the neural net. This network can be readily extended to allow for adaptability for different parameters of the dynamical system. Say that the system has some set of parameters $\lambda$, hence $\dot{\mathbf{x}} = f(\mathbf{x}; \lambda)$, we can then add additional input channels to the ODEnet for $\lambda$. The network then attempts to learn $f$ as a function of $\mathbf{x}$ and $\mathbf{\lambda}$. Adaptability may be possible by training the network on a wide range of $\lambda$ values, we demonstrate this for the Adaptable Hamiltonian Network in Sec. \ref{HNN}. \\

\begin{figure}
    \centering
    \includegraphics[width=0.5\textwidth]{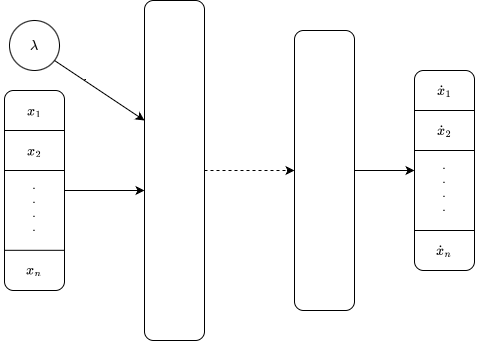}
    \caption{General Architecture of an Adaptable ODEnet}
    \label{fig:my_label}
\end{figure}

A more sophisticated method is to make the network recurrent, this is especially useful when the data is widely separated in time and finite differences may be inaccurate. This is accomplished by using the $\dot{\mathbf{x}}(t)$ outputted by the network along with an integrator to predict $\mathbf{x}(t+\Delta t)$. In general one can input a time series for $\mathbf{x}$ and predict the vector for an arbitrary number of steps. The Loss function is then defined between the predicted and true time series avoiding the use of finite differences for the the same.\\

As seen in this next section, modified ODENets form the basis for networks that obey Hamiltonian Dynamics.

\section{Encoding Hamiltonian Dynamics}

\begin{figure}
    \centering
    \subfloat[Architecture of an HNN \cite{Greydanus_Dzamba_Yosinski_2019}]{\includegraphics[width=0.45\textwidth]{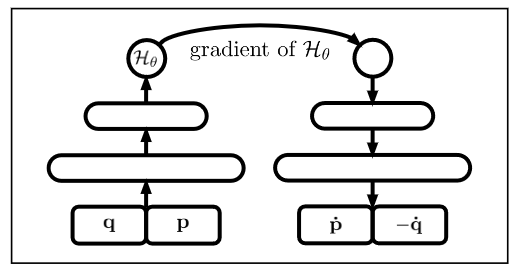} \label{fig:HNN}} 
    \subfloat[Adaptable Hamiltonian Neural Network \cite{Han_Glaz_Haile_Lai_2021}]{\includegraphics[width=0.45\textwidth]{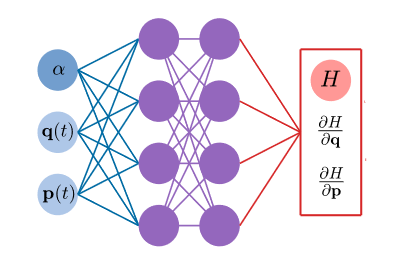} \label{fig:AHNN}} 
    \caption{Hamiltonian Neural Network}
    \label{fig:my_label}
\end{figure}

Generic Neural Networks(both feedforward and recurrent) suffer from the curse of overfitting, i.e. the model fits too closely to the training data such that it captures noise as well rather than the underlying patterns that generalize to new data. However, the goal of using NNs in physics is to be able to leverage their complexity and learning capabilities to discover underlying physical laws in domains where traditional methods fail. Hence we would ideally like to be able leverage any knowledge we have of the physical system such as symmetries, partial underlying equations, boundary conditions, etc. In this work, we aim to develop models that would obey known physical laws such as respecting the conserved quantities of the system. As mentioned earlier, Greynadus et al. \cite{Greydanus_Dzamba_Yosinski_2019} developed a neural network architecture(Fig. \ref{fig:HNN})  which leverages the laws of Hamiltonian Dynamics to model systems that obey the same but for which the Hamiltonian may not be known or expressible in closed form.

\subsection{Hamiltonian Neural Networks} \label{HNN}

In Hamilton's formulation of classical mechanics, the dynamical equations that define time evolution of the canonical variables $(q, p)$ are given by the partial derivatives of a function referred to as the Hamiltonian $\mathcal{H}(\mathbf{q}, \mathbf{p}, t)$. This function is defined by a Legendre Transformation of the Lagrangian($\mathcal{L}(\mathbf{q}, \dot{\mathbf{q}}, t) = T - V$):

\begin{equation}
    \mathcal{H}(\mathbf{q}, \mathbf{p}, t) = \mathbf{p} \cdot \mathbf{\dot{q}} - \mathcal{L}(\mathbf{q}, \dot{\mathbf{q}}, t)
\end{equation}
Where $p_i = \dfrac{\partial \mathcal{L}}{\partial \dot{q}_i}$ are the \textit{Canonical Momentum} variables. Hamilton's equations define the dynamical system:

\begin{gather}
    \dot{\mathbf{q}} = \frac{\partial \mathcal{H}}{\partial \mathbf{p}} \\
    \dot{\mathbf{p}} = -  \frac{\partial \mathcal{H}}{\partial \mathbf{q}}
\end{gather}

We consider only dynamical systems for which $\mathcal{H} = \mathcal{H}(\mathbf{q}, \mathbf{p})$, i.e. the Hamiltonian is not explicitly dependent on time and is a conserved quantity. \\

To model systems which obey Hamiltonian Dynamics(i.e. Hamilton's equations) for which we may not know the closed form of the Hamiltonian, we construct a Neural Network that takes in canonical variables $(\mathbf{q}, \mathbf{p})$ as input and has only a single neuron as output that we intend returns $\mathcal{H}(\mathbf{q}, \mathbf{p})$. Hence we aim to learn the Hamiltonian as a function of canonical variables with some NN parameters $\theta$ : $\mathcal{H}_{\theta} (\mathbf{q}, \mathbf{p})$. In order to enforce Hamilton's equations, we take the gradient of $\mathcal{H}_{\theta}$ w.r.t the inputs $(\mathbf{q}, \mathbf{p})$. The loss function is then defined by:

\begin{equation}
    \mathcal{L}_{(\mathbf{q}, \mathbf{p})}(\theta) = \norm{\frac{\partial H_{\theta}}{\partial \mathbf{p}} - \frac{\partial \mathbf{q}}{\partial t}}_2 + \norm{\frac{\partial H_{\theta}}{\partial \mathbf{q}} + \frac{\partial \mathbf{p}}{\partial t}}_2  
\end{equation}

This way, by minimising the loss, we ensure that the $\mathcal{H}_{\theta} (\mathbf{q}, \mathbf{p})$ that the network has learnt is the Hamiltonian of the system or a scaled version of it. This works similar to an ODENet in the sense that the loss is a function of $(\dot{\mathbf{q}}, \dot{\mathbf{p}})$ for which we can either use finite time differences or, for illustration purposes, analytical time derivatives(if the equations are known). \\

Once trained this network can be used along with an integrator to predict the time series for arbitrary durations.

\subsubsection{Adaptability}

Analogous to how we modified ODENets to allow for adaptability, one can add parameter channels to HNN input so that a Hamiltonian $\mathcal{H}(\mathbf{q}, \mathbf{p}; \alpha)$ can be modelled, where $\alpha$ represents the parameters(Fig. \ref{fig:AHNN}). Han et al. \cite{Han_Glaz_Haile_Lai_2021} found that Adapatable HNNs(AHNNs) trained on a sufficiently large amount of different parameter trajectories, are capable of interpolating between the training parameter values i.e. it is able to predict dynamics and parameter points that it is not exposed to during training.


\subsection{Adaptable Symplectic Recurrent Neural Networks(ASRNNs)}

Similar to how we extended ODENets to recurrent versions of themselves, one can do the same for HNNs with the use of a suitable integrator. To ensure that the time evolution predicted by the recurrent network obeys the structure of Hamilton's equations, it is necessary to make use of \textbf{symplectic integrators}. These are designed to perform the numerical evolution of Hamiltonian systems and preserve the \textit{symplectic} structure of the system, which is a property that ensures the conservation of energy and the preservation of important geometric properties such as phase space volume(Liouville's theorem). These integrators hence ensure that error in conserved quantities such as energy remains bounded and does not grow with integration time. One can hence construct an RNN architecture that implements the leapfrog algorithm with a HNN used to provide the time derivatives $\dot{\mathbf{p}}$ and $\dot{\mathbf{q}}$. This was first proposed by Chen et al. \cite{Chen_Zhang_Arjovsky_Bottou_2020}. To allow ourselves to take advantage of Symplectic integrators we restrict our attention to systems with \textit{separable} Hamiltonians, i.e. $\mathcal{H}(\mathbf{q}, \mathbf{p}) = \mathcal{K}(\mathbf{p}) + \mathcal{V}(\mathbf{q})$. One can then construct Hamiltonian networks for $\mathcal{K}$ and $\mathcal{V}$ separately s.t. $\mathcal{H_{\theta}(\mathbf{q}, \mathbf{p})} = \mathcal{K}_{\theta_1}(\mathbf{p}) + \mathcal{V}_{\theta_2}(\mathbf{q})$.

The integration algorithm is as follows:

\begin{gather}
    \mathbf{p}\left( t+\frac{\Delta t}{2}\right) = \mathbf{p}(t) - \frac{\Delta t}{2} \frac{\partial \mathcal{V}_{\theta_2}}{\partial \mathbf{q}} \bigg|_t \\
    \mathbf{q}\left( t+\Delta t\right) = \mathbf{q}(t) + \Delta t \frac{\partial \mathcal{K}_{\theta_1}}{\partial \mathbf{p}} \bigg|_{t+\frac{\Delta t}{2}} \\
    \mathbf{p}\left( t+\Delta t\right) =  \mathbf{p}\left( t+\frac{\Delta t}{2}\right) - \frac{\Delta t}{2} \frac{\partial \mathcal{V}_{\theta_2}}{\partial \mathbf{q}} \bigg|_{t+\Delta t}
\end{gather}

Hence the symplectic recurrent neural network(SRNN) takes in the initial state $(\mathbf{q}_0, \mathbf{p_0})$ and predicts an arbitrary number of steps with some time step. The loss function is defined by:

\begin{equation}
    \mathcal{L}_{(\mathbf{q}_0, \mathbf{p_0})} (\theta) = \sum_t \left(\norm{\mathbf{q}_t - \mathbf{\hat{q}_t}}_2 + \norm{\mathbf{p}_t - \mathbf{\hat{p}_t}}_2  \right)
    \label{SRNNloss}
\end{equation}

Where $(\mathbf{\hat{q}}, \mathbf{\hat{p}})$ is the ground truth data and $(\mathbf{q},\mathbf{p})$ are the NN outputs. During training the series length can be fixed or variable: if we fix it to $1$ we define the network to be \textit{single step} trained. When predicting, the series length can be arbitrarily long and due to its symplectic structure, errors in conserved quantities will not grow over time. \\

Chen et al. \cite{Chen_Zhang_Arjovsky_Bottou_2020} found that SRNNs significantly outperform simple Hamiltonian Neural Networks in the analysis of more complex potentials and in terms of long term prediction. Further HNNs require time derivatives of the canonical coordinates for training, these are can either be calculated from analytical expressions(if available) or through the use of finite differences of the time series. The latter can be very inaccurate especially in conditions where the data is widely separated in time.
SRNNs mitigate this problem by making use of only the elements of the time series of $(\mathbf{q}, \mathbf{p})$ in their loss function. We propose a novel architecture that augments these SRNNs with additional parameter channels to allow for adaptability: this is done by making use of AHNNs instead of simple HNNs in the implementation of the leapfrog algorithm. Assuming for simplicity that the parameters are purely fed into the potential function(this can always be done with the choice of appropriate coordinates), it is straightforward to extend the formulation. We hence now aim to model a potential $\mathcal{V} = \mathcal{V}(q; \lambda)$, the Hamiltonian we learn can be represented as:
\begin{equation}
    \mathcal{H}_{\theta}(\mathbf{q}, \mathbf{p} ; \lambda) = \mathcal{K}_{\theta_1}(\mathbf{p}) + \mathcal{V}_{\theta_2}(\mathbf{q}; \lambda)
\end{equation}

\begin{figure}[ht]
    \centering
    \includegraphics[width=0.5\textwidth]{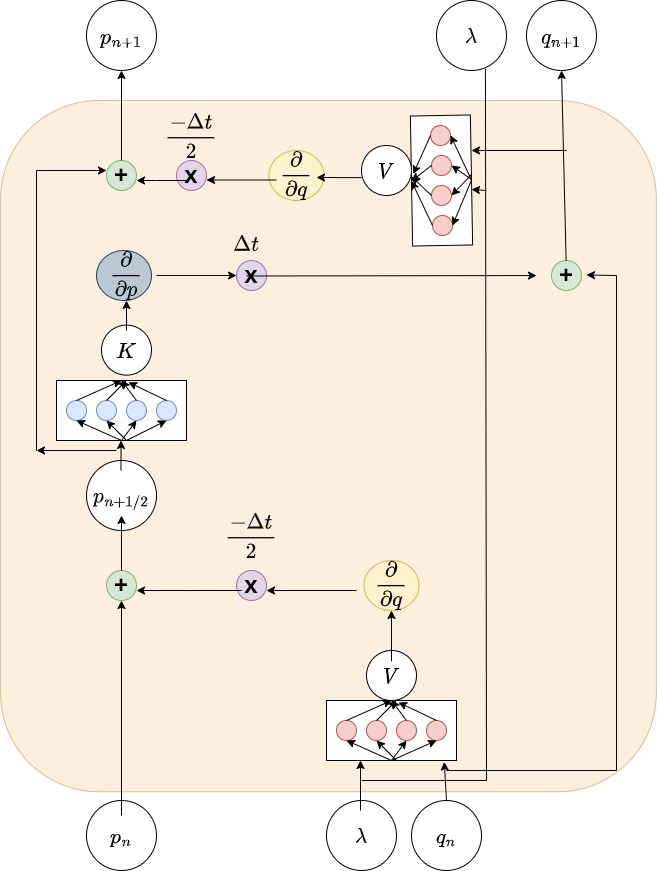}
    \caption{Adaptable Symplectic Recurrent Neural Network}
    \label{fig:symprnn}
\end{figure}

The $\lambda$ remains constant in time hence the leapfrog algorithm update equations remain as before, our proposed architecture is illustrated in Fig. \ref{fig:symprnn}.

These networks retain the loss function defined by Eqn. \ref{SRNNloss} and hence the algorithm for training remains unchanged. However, the input data can now contain trajectories corresponding to different values of $\lambda$, hence the network  incorporates this information in its structure when attempting to discover the Hamiltonian. We refer to this architecture as an \textbf{Adaptable Symplectic Recurrent Neural Network} \footnote{A pytorch implementation of this architecture can be found \href{https://github.com/Vedanta-T/Adaptable-Sympletic-Recurrent-Neural-Networks}{here}} (Fig. \ref{fig:symprnn}). As illustrated in the following chapter, these networks are demonstrated to be capable of learning the entire parameter space given the dynamical information at only a few points. 

\section{Illustration : The Linear Harmonic Oscillator}

\begin{figure}[ht]
    \centering
    \includegraphics[width=0.8\textwidth]{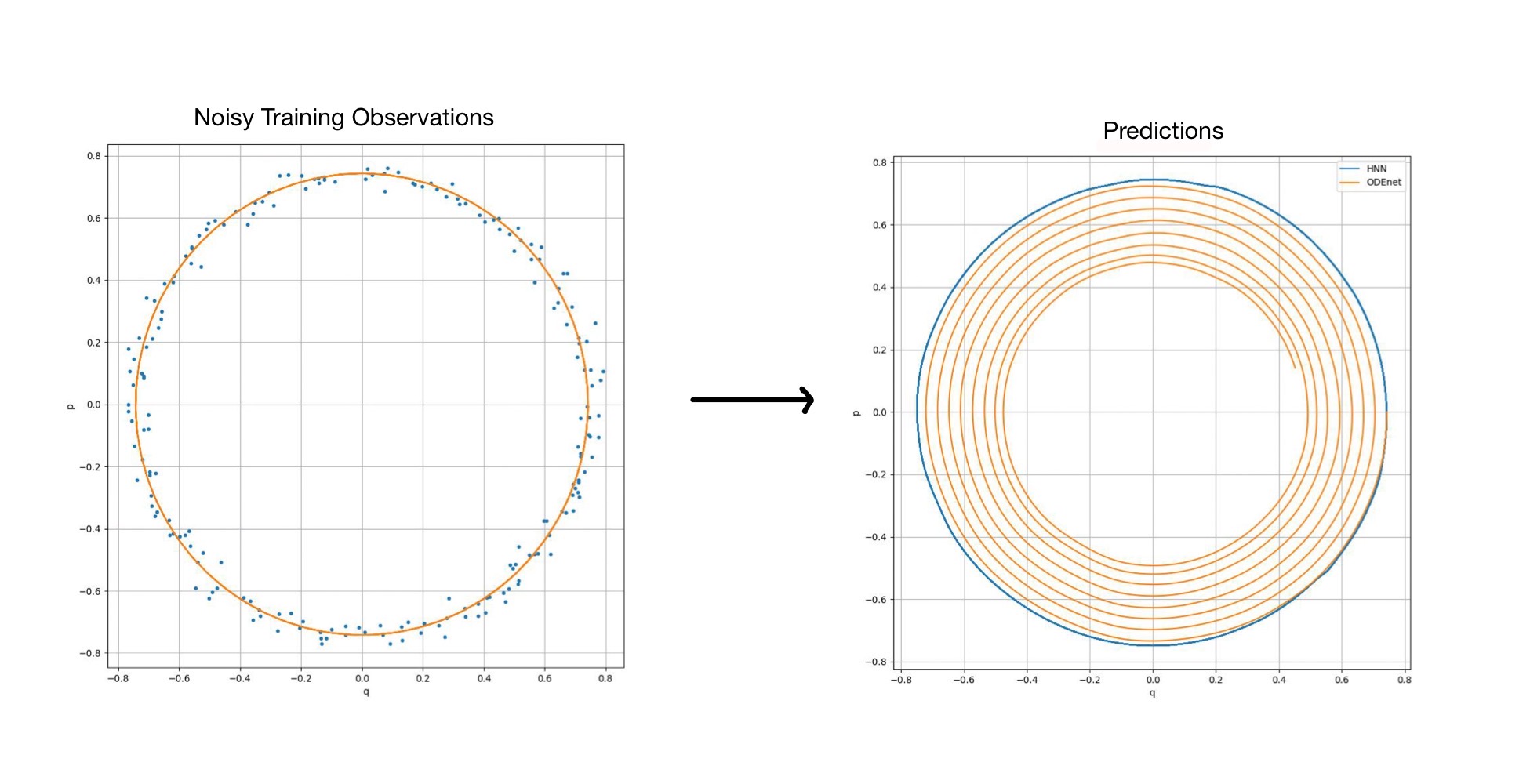}
    \caption{Comparison of Long term prediction with HNNs and ODEnets}
    \label{fig:oscillator1}
\end{figure}

One can compare the effectiveness of Physics Informed Machine learning over traditional Deep learning methods through systems as simple as the Linear Harmonic Oscillator. Fig. \ref{fig:oscillator1} shows the performance of an HNN over a simple feedforward neural network: as can be seen the HNN is able to maintain closed orbits without degradation for long periods of time even with training from noisy data. \\

\begin{figure}[ht]
    \centering
    \subfloat[Energy Conservation]{\includegraphics[width=0.45\textwidth]{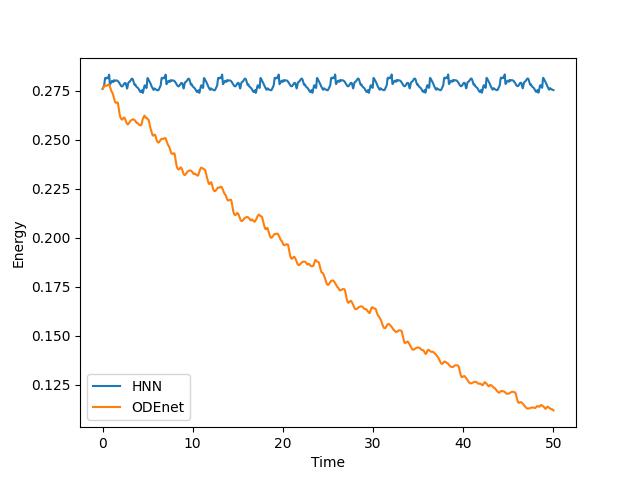}}
    \subfloat[HNN Conserved Quantity]{\includegraphics[width=0.45\textwidth]{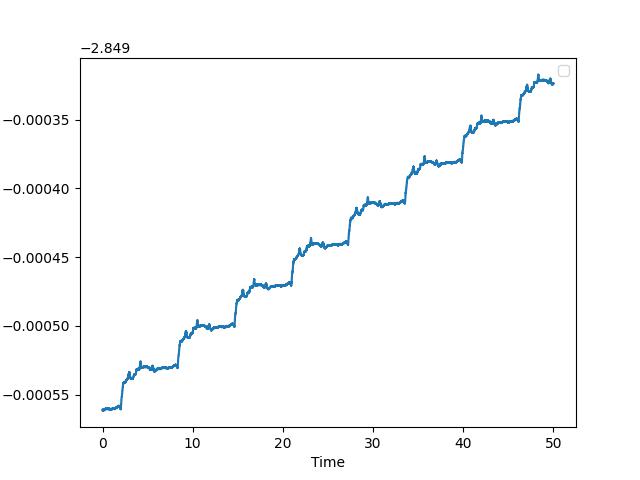}}
    \caption{Conserved Quantities for the Harmonic Oscillator}
    \label{fig:HOenergy}
\end{figure}

Further this implies the network respects conservation of energy to a far greater degree as well. Fig. \ref{fig:HOenergy} shows its energy conservation characteristics as well as the conservation of the `HNN conserved quantity' which is similar to the total energy but not exactly the same. We explain this further in the following chapter \\

It is clear that Hamiltonian Neural networks outperform simple ODEnets with even the simplest of Hamiltonian systems by preserving the conservation of energy to a great degree. This behaviour is reflected if one considers the inclusion of a additional parameter channel as well. However in the case of systems with multiple input parameter channels, both HNNs and ODEnets struggle to learn the dynamics. In this case we turn to the Symplectic Recurrent networks, specifically the \textbf{Adaptable Symplectic Recurrent Neural Networks} which we propose.

To summarize, it was noted by Greynadus et al. \cite{Greydanus_Dzamba_Yosinski_2019} that simple HNNs, although outperforming ODEnets, failed to generalize to high dimensional nonlinear systems such as the 3 body problem, Chen et al. \cite{Chen_Zhang_Arjovsky_Bottou_2020} showed their architecture was able to learn the dynamics of the 3 body problem, however was not adaptable in nature. Han et al. \cite{Han_Glaz_Haile_Lai_2021} added input parameter channels to HNNs and their architecture showed great promise with Hamiltonians restricted to a single varying parameter. However, under multiparameter conditions this method failed to effectively incorporate the parameter information and learn the Hamiltonian dynamics of the system under consideration.

In the following chapter we study the robustness of our proposed architecture(\textbf{ASRNN}), a combination of the methods mentioned in the previous paragraph, and investigate whether its able to tackle some of the challenges faced by these existing networks.

\chapter{Robustness of ASRNNs} \label{chapter3}
This dissertation attempts to study two primary questions: 
\begin{itemize}
    \item Does the newly proposed architecture of an Adaptable Symplectic Recurrent Neural Network outperform the existing Adaptable Hamiltonian Networks in learning complex potentials and forecasting long term dynamics.
    \item Using the variety of methods discussed in the previous Chapter,  can we use the high dimensional and nonlinear capabilities of Neural Networks to learn the dynamics of Hamiltonian systems given only partial information. (Chapter \ref{chapter4})
\end{itemize}

For our analysis we require a sufficiently complex Hamiltonian which allows for great diversity in its dynamics while also being relatively low dimensional to permit analysis of it with our limited computing resources. We would also prefer to have non-linearities to allow for chaotic motion and for complex attractors to form; additionally most real world systems we would like to apply the formulation on are nonlinear. Keeping the above in mind we chose to work with the Henon-Heiles Hamiltonian \cite{HenonHeiles} defined in its most general form by:

\begin{equation}
    \mathcal{H}(q_x, q_y, p_x, p_y; m, \omega_x, \omega_y, \alpha, \beta) = \frac{p_x^2 + p_y^2}{2m} + \frac{1}{2}m \left(\omega_x^2 q_x^2 + \omega_y^2 q_y^2\right) + \alpha q_x^2 q_y - \beta \frac{q_y^3}{3}
\end{equation}

This potential was proposed by French mathematicians Michel Henon and Carl Heiles as a toy model for dynamics of stars in galaxies, which are influenced by gravitational interactions between each other. It is constructed in 2 dimensional configuration space(4 dimensional phase space) with two perpendicular linear harmonic oscillators coupled through an asymetric non linear perturbation. Henon and Heiles aimed to produce the simplest model for stellar motion in galaxies which is typically modelled through a sum of perpendicular linear harmonic oscillators. The non-linear perturbation allows for richer dynamics including chaotic, quasiperiodic, resonant, bifurcations and periodic motion. \\

For our analysis we consider $m = \omega_x = \omega_y = 1$, i.e. we are working in units where mass of the body under the potential and the frequency of the oscillators is normalized to one. The Hamiltonian then reduces to:

\begin{equation}
    \mathcal{H}(q_x, q_y, p_x, p_y; \alpha, \beta) = \frac{p_x^2 + p_y^2}{2} + \frac{q_x^2 + q_y^2}{2} + \alpha q_x^2 q_y - \beta \frac{q_y^3}{3}
\end{equation}

Through this chapter we'll consider several combinations of $\alpha$ and $\beta$, for reference the standard Henon-Heiles is taken as the case for which $\alpha = \beta = 1$.

\section{Single Parameter Conditions}

Lets first consider only the single parameter Henon-Heiles, i.e. $\beta = \alpha$, hence $\alpha$ alone scales the nonlinearity. Let $\mathbf{q} = \begin{pmatrix} q_x \\ q_y \end{pmatrix}$, $\mathbf{p} = \begin{pmatrix} p_x \\ p_y \end{pmatrix}$, the Hamiltonian and equations of motion are given by:\\

\begin{equation}
    \mathcal{H}(\mathbf{q}, \mathbf{p}; \alpha) = \frac{p_x^2 + p_y^2}{2} + \frac{q_x^2 + q_y^2}{2} + \alpha \left(q_x^2 q_y - \frac{q_y^3}{3}\right)
\end{equation}

And its associated equations of motion:
\begin{gather}
    \dot{\mathbf{q}} =  \mathbf{p} \\
    \dot{\mathbf{p}} = \begin{pmatrix}
                       -q_x - 2\alpha q_xq_y \\
                       -q_y - \alpha(q_x^2 - q_y^2)
                       \end{pmatrix}
\end{gather}

\begin{figure}[ht]
    \centering
    \subfloat[]{\includegraphics[width=0.5\textwidth]{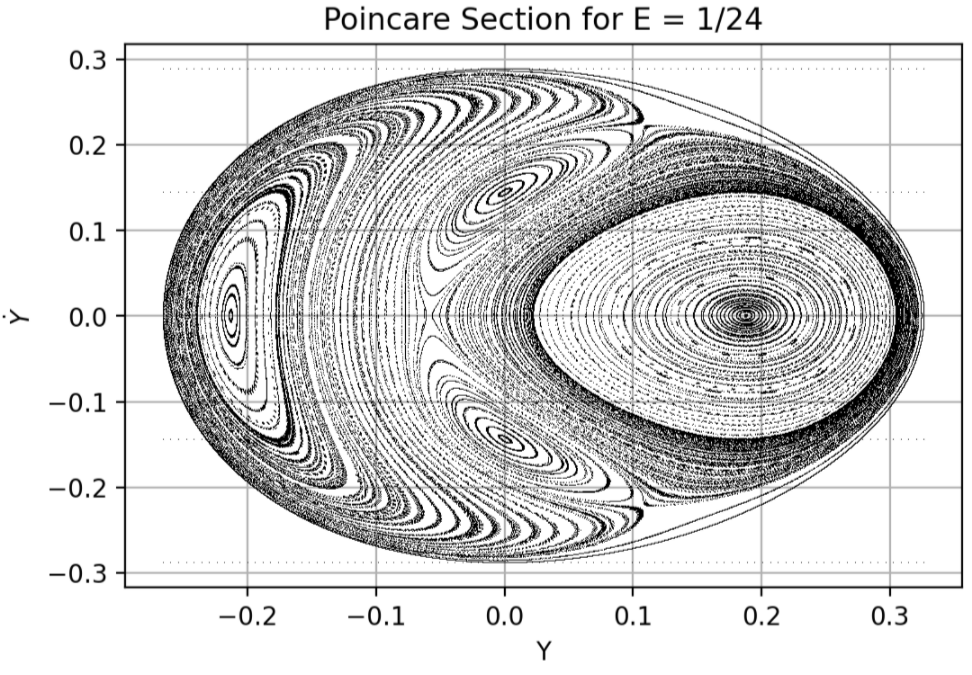}}
    \subfloat[]{\includegraphics[width=0.5\textwidth]{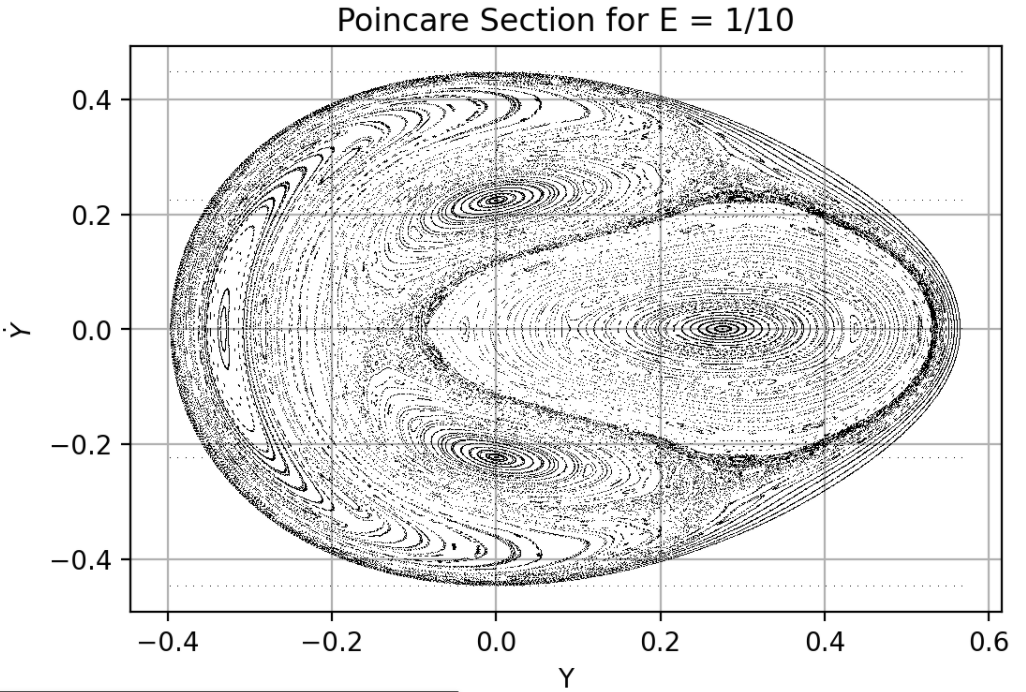}} \\
    \subfloat[]{\includegraphics[width=0.5\textwidth]{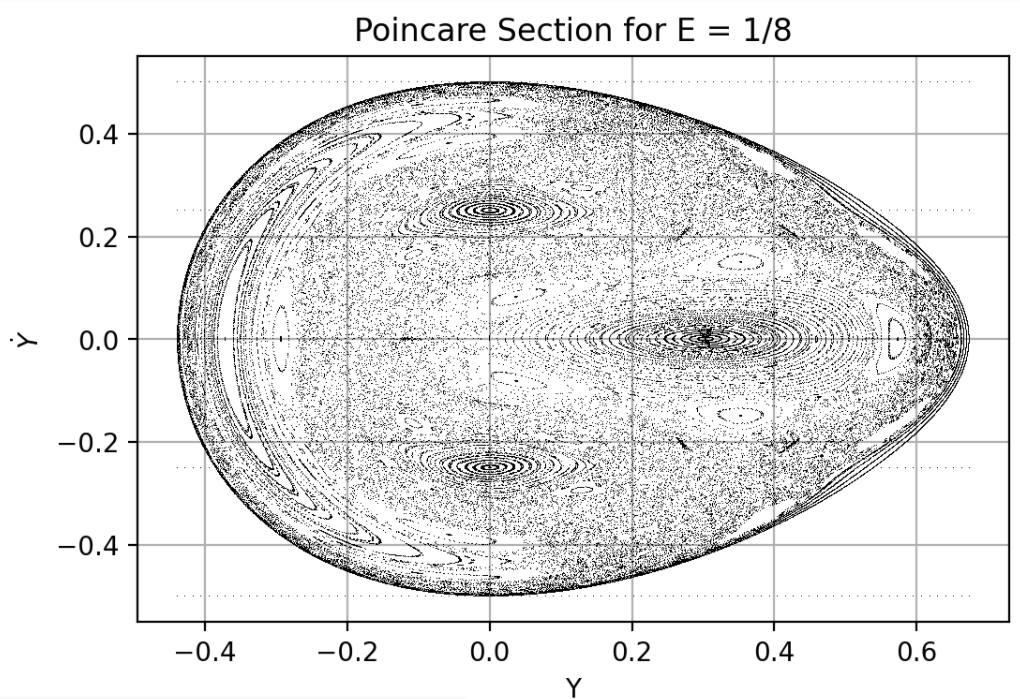}} 
    \subfloat[]{\includegraphics[width=0.5\textwidth]{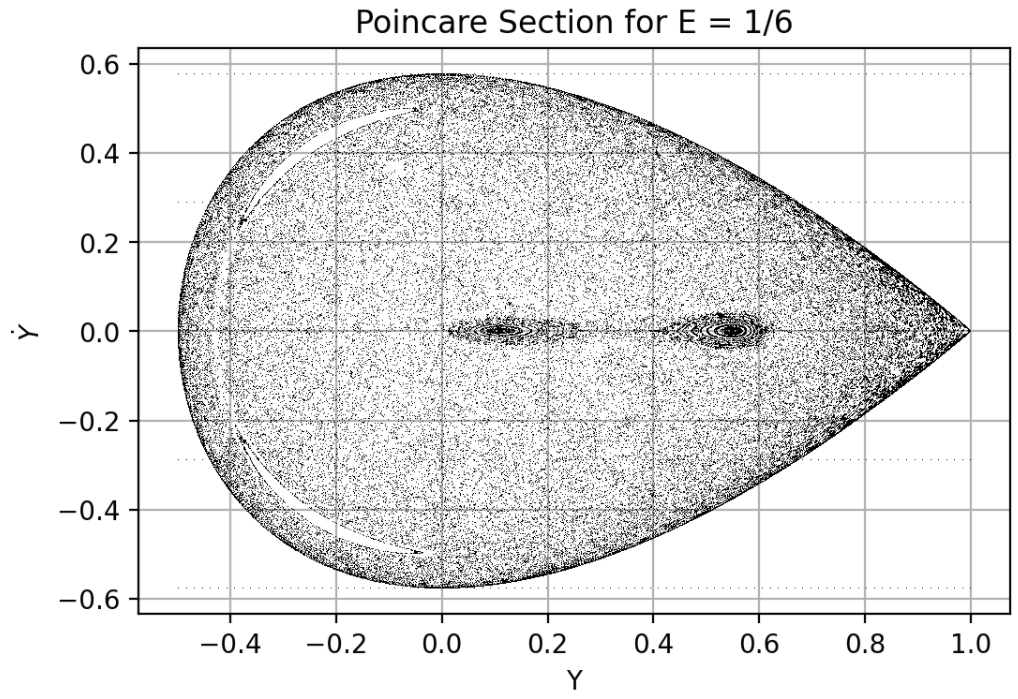}}
    \caption{Poincare Sections for $\alpha = 1$ sliced parallel to the $(q_y, p_y)$ plane}
    \label{fig:poincare}
\end{figure}

Note the $q_xq_y$ term, its mixed nature allows for the great complexity in dynamics that the Henon Heiles displays despite its relatively simple appearance. 
It is well established that for $0 < \alpha \leq 1$, the potential has a finite escape energy : orbits for $E\leq\frac{1}{6}$ remain bounded for all initial conditions. For $\alpha<1$ the escape energy is usually higher. Furthermore for low enough $\alpha$ almost all orbits are periodic or quasiperiodic, however as the strength of the perturbation is increased, some invariant tori survive while several are deformed leading to chaotic motion. For energies close to the escape energies most orbits are chaotic, Fig. \ref{fig:poincare} shows the Poincare sections for some initial energies at $\alpha=1$, these sections correspond to the $(q_y, p_y)$ plane. Closed curves represent stable periodic orbits while irregular broken ones represent quasiperiodic or chaotic motion. It is clear that at higher energies the stable orbits are largely broken with only a few `islands of stability' remaining. 

Thus the Henon-Heiles even with a single parameter allows for a very diverse range of dynamical behaviour, we hence analyse the potential both in single parameter and multiparameter configurations. The rest of this chapter demonstrates the capability of Adaptable Symplectic Neural Networks as a robust extension of AHNNs while the following explores their coupling with LSTMs.

\subsection{Training an AHNN} \label{single parameter AHNN}

We independently train an AHNN and ASRNN for the same data, formatted as needed for the network architecture. Data is generated under the following conditions:

\begin{itemize}
    \item A discrete set of 4 parameter values is considered $\alpha \in \{0.2, 0.4, 0.6, 0.8\}$. For each of these values $50$ initial conditions for each of the following values of energy are taken: $E_{in} \in \{\frac{1}{24}, \frac{1}{12}, \frac{1}{8}, \frac{1}{6}\}$.
    \item For this total of $200$ initial conditions, time series $(\mathbf{q}, \mathbf{p}, \dot{\mathbf{q}}, \dot{\mathbf{p}})$ of length $300$ are constructed using the Leapfrog algorithm with a time step of $0.001$. This data is coarse grained for subsequent points in the time series to have separation $0.1$.
    \item Finally we are left with $200$ series of length $3000$ data points each. Removing a transient of $500$ data points the total number of $(\mathbf{q}, \mathbf{p}, \dot{\mathbf{q}}, \dot{\mathbf{p}})$ data points is $500,000$.
    \item Note that each $(\mathbf{q}, \mathbf{p}, \dot{\mathbf{q}}, \dot{\mathbf{p}})$ is treated as a separate training data point. The total data is divided into training and validation sets and training is performed as described in the previous chapter
\end{itemize}

An adaptable HNN is constructed with the following architecture as done by \cite{Han_Glaz_Haile_Lai_2021}:

\begin{itemize}
    \item The inputs to the AHNN are the pairs $(\mathbf{q}, \mathbf{p})$ with an additional parameter channel for $\alpha$. The outputs meanwhile are the corresponding $(\dot{\mathbf{q}}, \dot{\mathbf{p}})$ pairs obtained from the network by automatic differentiation.
    \item The network has two hidden layers of $200$ neurons each with activation function $Tanh$.
    \item The network is trained for $500$ epochs with a batchsize of $128$. Pytorch is used to implement $SGD$ training with an \textbf{Adam} optimizer \cite{Kingma2014Dec}. 
\end{itemize}

The network trains and performs well not only on the training $\alpha$ values but $\forall \ \alpha \in [0.2, 0.8]$ demonstrating great adaptability. This verifies the results obtained by \cite{Han_Glaz_Haile_Lai_2021}. However it should be noted that the AHNN is not robust in terms of long term predictions and its performance degrades sharply outside the training parameter space. We refer to this further in Sec. \ref{Comparison}.

\subsection{Training the ASRNN} \label{single parameter SRNN}

An ASRNN can be trained in two modes: single step and recurrent. The former implies that the Leapfrog algorithm takes only a single step and the loss function takes in simply that predicted step. In case of the latter, we predict multiple steps starting from the given initial condition and the loss function is a superposition of that for each step. One may choose one or the other depending on the problem at hand, in this chapter we use recurrent training which Chen et al. \cite{Chen_Zhang_Arjovsky_Bottou_2020} find to be more resistant to noise and can discern longer trajectories. Although we do not go into it in this work, an independent analysis of the effect of step size is needed to optimize these networks further. \\

We now return to training the ASRNN, the same data is used as in the previous section however its formatting is different. It is processed as follows:
\begin{itemize}
    \item Data for only $(\mathbf{q}, \mathbf{p})$ is used for training an ASRNN, analytical time derivatives are not needed at all.
    \item The longer time series are divided into shorter sections, specifically sections of length $11$ where the first point is the initial state provided to the ASRNN, and the following 10 steps are predicted.
\end{itemize}

The ASRNN is constructed with a single hidden layer of $2048$ neurons which uses the $Tanh$ activation function, trained for $500$ epochs with the Adam Optimizer. The ASRNN not only performs better than the AHNN in predicting trajectories in the training parameter interval, it further shows the capability to learn chaotic dynamics in the parameter regime $[0.8, 1]$. 

\subsection{Comparison of Results} \label{Comparison}

As our primary measure, we define relative energy error as:
\begin{equation}
    \Delta E(t) = \frac{|E_{net}(t) - E_{true}(t)|}{E_{true}(t)} \times 100  \ \%
\end{equation}
Where $E_{true}(t)$ represents the energy of the ground truth trajectory integrated using analytical derivatives with the Leapfrog algorithm($\Delta t = 0.001$) at time $t$. $E_{net}(t)$ is the energy of the neural network predicted trajectory at time $t$. The mean relative energy error $\langle \Delta E \rangle$ is then defined as the average of $\Delta E(t)$ over the entire trajectory. \\

\begin{figure}
    \centering
    \includegraphics[width=\textwidth]{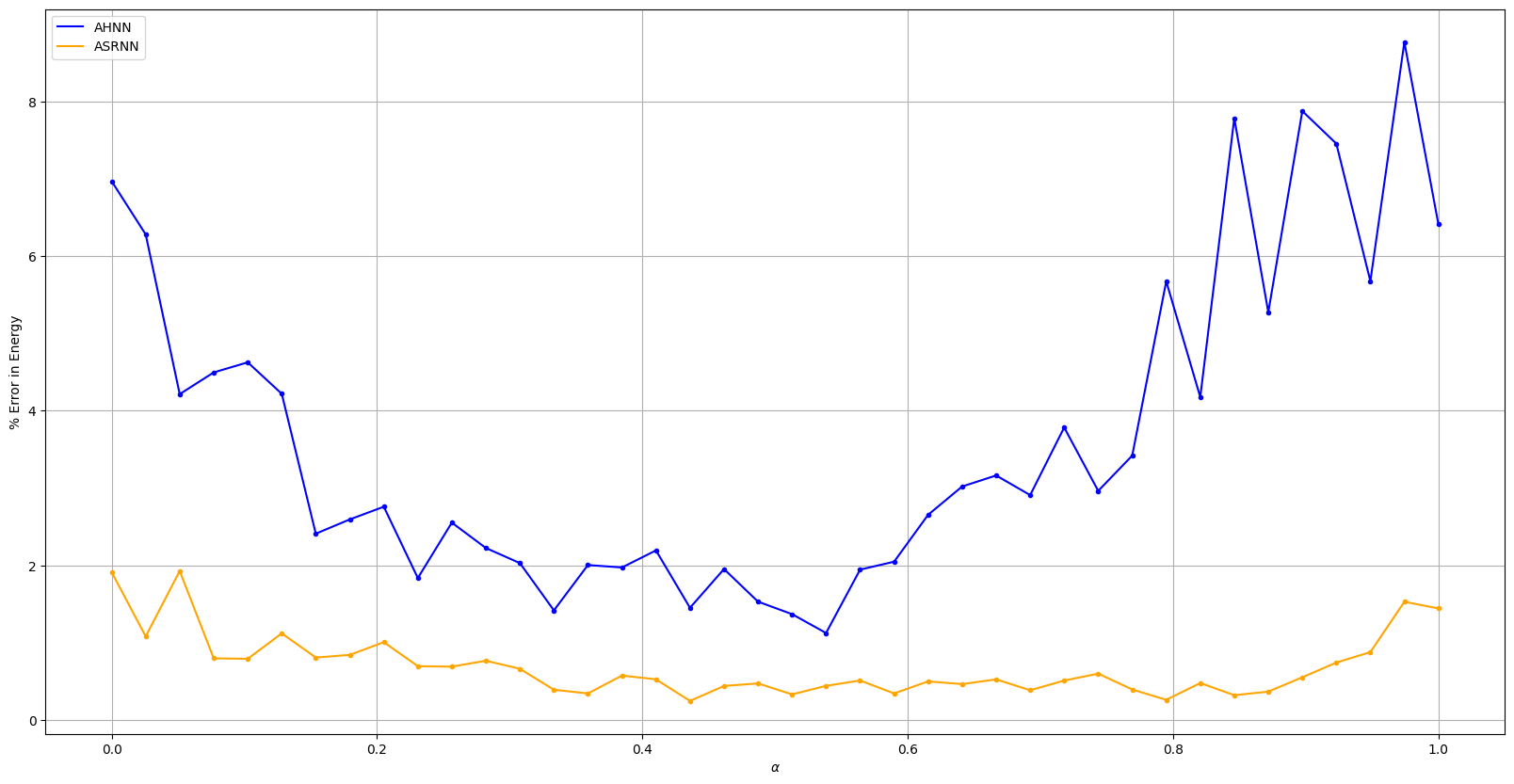}
    \caption{$\langle \Delta E \rangle$ for $\alpha \in (0, 1]$ averaged over several initial conditions}
    \label{fig:singleparamdeltaE}
\end{figure}

Within the training region both the ASRNN and AHNN perform similarly within the training parameter regime $[0.2,0.8]$ though the ASRNN outperforms the AHNN. However out of this region the mean relative potential energy error of the AHNN predicted trajectories rises sharply to over $5\%$ as reported by Han et al. \cite{Han_Glaz_Haile_Lai_2021}. If one speaks of the mean relative error in energy the results are even worse with the quantity rising to over $8\%$. In the unstable high energy space $E = \frac{1}{6}$ with $\alpha \in [0.8, 1]$ some initial conditions lead to unbounded trajectories, however if one considers an ensemble of independently trained networks this issue can be avoided. On the other hand, the ASRNN predicts trajectories with $\langle \Delta E \rangle$ consistently below $2\%$ over the entire parameter and phase space(Fig. \ref{fig:singleparamdeltaE}). It should also be noted that prediction with the ASRNN is significantly faster than the AHNN. \\

\begin{figure}
    \centering
    \includegraphics[width=0.5\textwidth]{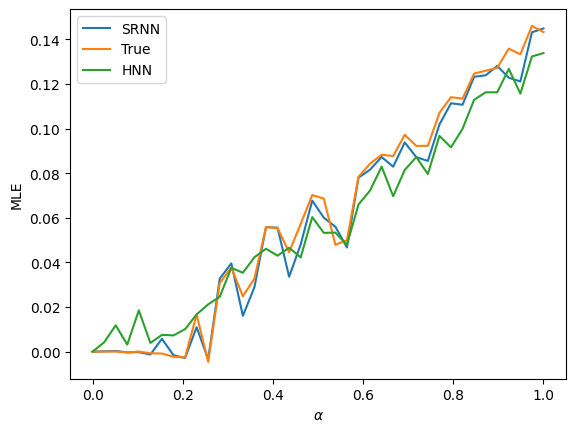}
    \caption{Lyapunov Exponent at High Energies}
    \label{fig:lyapsingle}
\end{figure}

In terms of long term prediction, $\Delta E$ remains bounded for ASRNNs due to the symplectic structure of the network, this is however not seen for AHNNs. The invariant tori of the Henon Heiles Hamiltonian break down at higher energies and parameter values, hence, one sees a transition to chaos fingerprinted by the rising Lyapunov exponent(Appendix \ref{lyapunovalgorithm}): a measure of the divergence of nearby trajectories in phase space. We can demonstrate that both ASRNNs and AHNNs retain these properties of the system as seen in the similarity of the Lyapunov exponent calculations for both of them alongside the ground truth trajectories(Fig. \ref{fig:lyapsingle})

\section{Multiparameter Conditions}

As discussed in the previous section, the ASRNN outperforms the AHNN performs even under single parameter conditions. Under multiparameter conditions this difference is even more pronounced. We now consider the case where $\alpha, \beta \in (0, 1]$. Data of the same number of batches and length of time as before is generated for all combinations of $(\alpha, \beta) \in \{0.2, 0.4, 0.6, 0.8\}$. The AHNN is trained with analytical time derivatives whereas the ASRNN is trained purely with the time series. Several measures are then used to test the performance of ASRNNs vs AHNN. \\

\begin{figure}
    \centering
    \includegraphics[width=\textwidth]{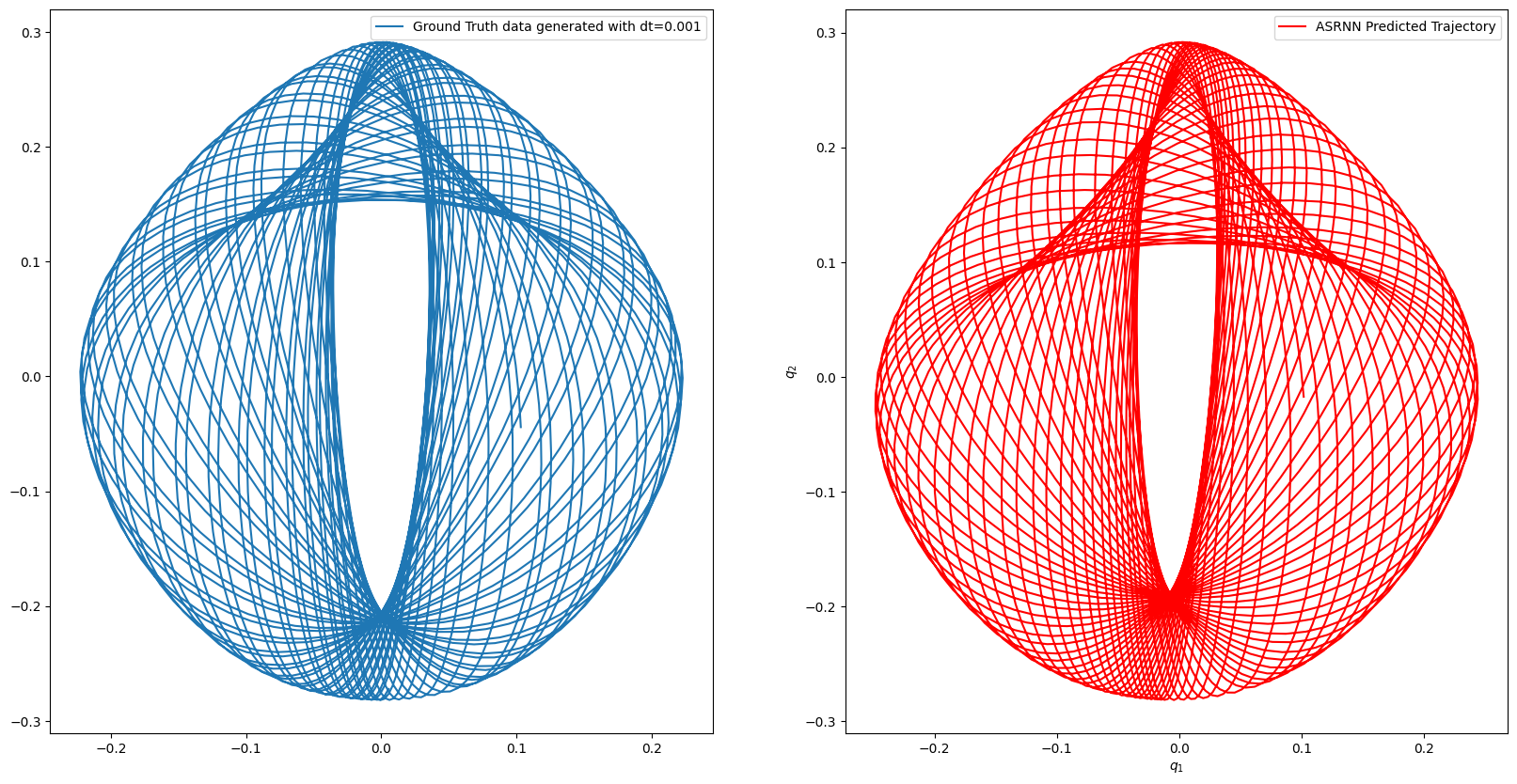}
    \caption{Predicted and true trajectories for $\begin{pmatrix}
        \alpha \\ \beta 
    \end{pmatrix} = \begin{pmatrix}
        1 \\ 0.2
    \end{pmatrix}$ with $E = \frac{1}{24}$ }
    \label{fig:1_0.2lowenergy}
\end{figure}

We first display some trajectories predicted by the ASRNN in comparison to those predicted with analytical time derivatives and the Leapfrog Algorithm to demonstrate the accuracy with which ASRNNs predict trajectories over long durations. Fig. \ref{fig:1_0.2lowenergy} shows an example for a quasiperiodic trajectory at low energies, we find the error in energy is of the order of $\sim 10^{-3}$. Even at higher energies, the ASRNN performs extremely well with energy conservation of the same order. One notes that even at parameter points which the net has not been trained on, the trajectories are of as close accuracy as the former: Fig. \ref{fig:0.9_0.7highenergy} shows an example at high energies. It should be noted this is a chaotic trajectory hence statistical properties should be used for comparison. As we saw in the previous section, the ASRNN was capable of mantaining great accuracy even for parameters outside its training parameter interval $[0.2, 0.8]$, this holds up even under multiparameter conditions as seen in the representative figures(Figs. \ref{fig:1_0.2lowenergy}-\ref{fig:0.9_0.7highenergy}). \\

\begin{figure}
    \centering
    \includegraphics[width=\textwidth]{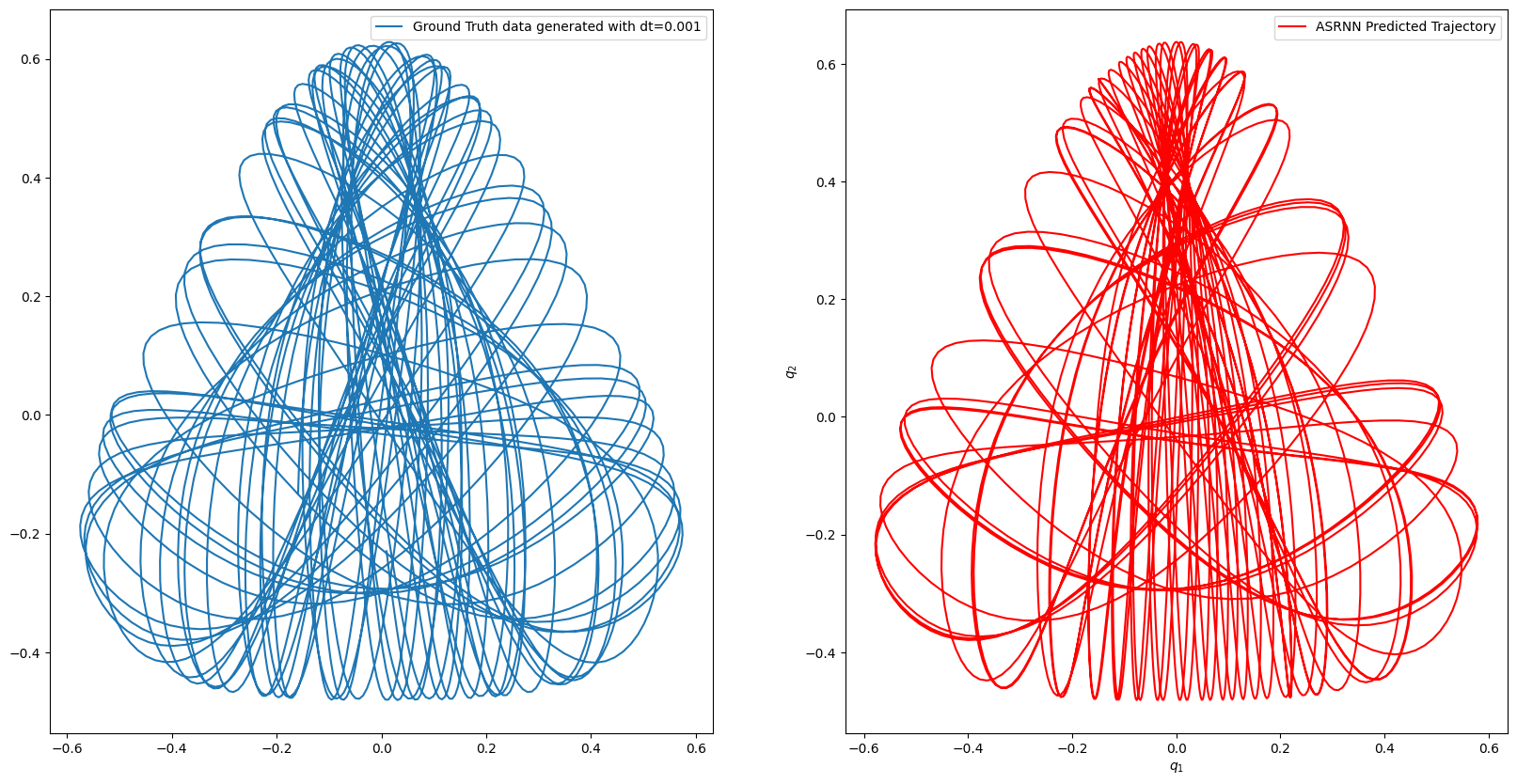}
    \caption{Predicted and true trajectories for $\begin{pmatrix}
        \alpha \\ \beta 
    \end{pmatrix} = \begin{pmatrix}
        0.9 \\ 0.7
    \end{pmatrix}$ with $E = \frac{1}{7}$}
    \label{fig:0.9_0.7highenergy}
\end{figure}

\begin{figure}
    \centering
    \includegraphics[width=\textwidth]{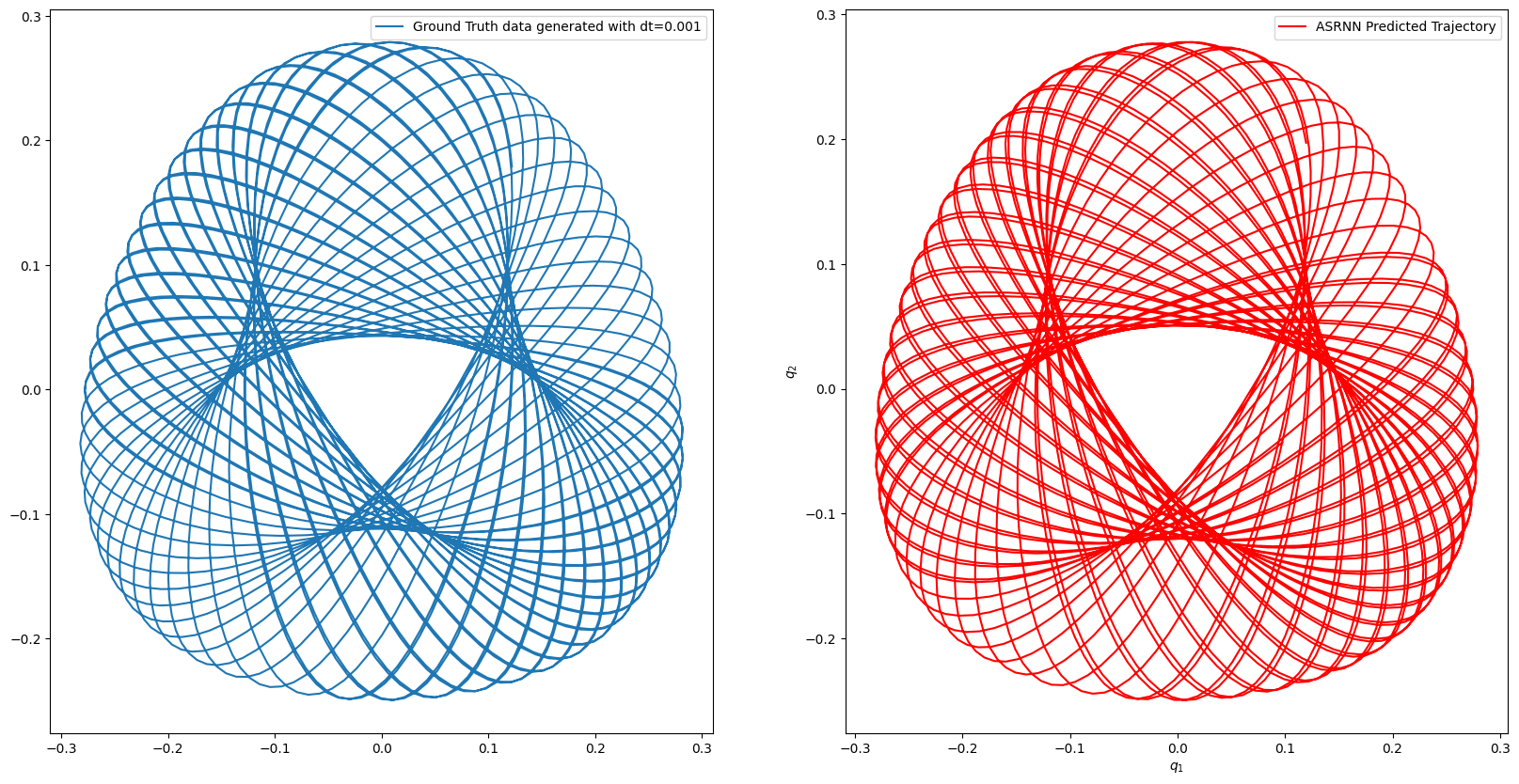}
    \caption{Predicted and true trajectories for $\begin{pmatrix}
        \alpha \\ \beta 
    \end{pmatrix} = \begin{pmatrix}
        0.7 \\ 0.9
    \end{pmatrix}$ with $E = \frac{1}{24}$}
    \label{fig:0.9_0.7highenergy}
\end{figure}

\begin{figure}
    \centering
    \includegraphics[width=\textwidth]{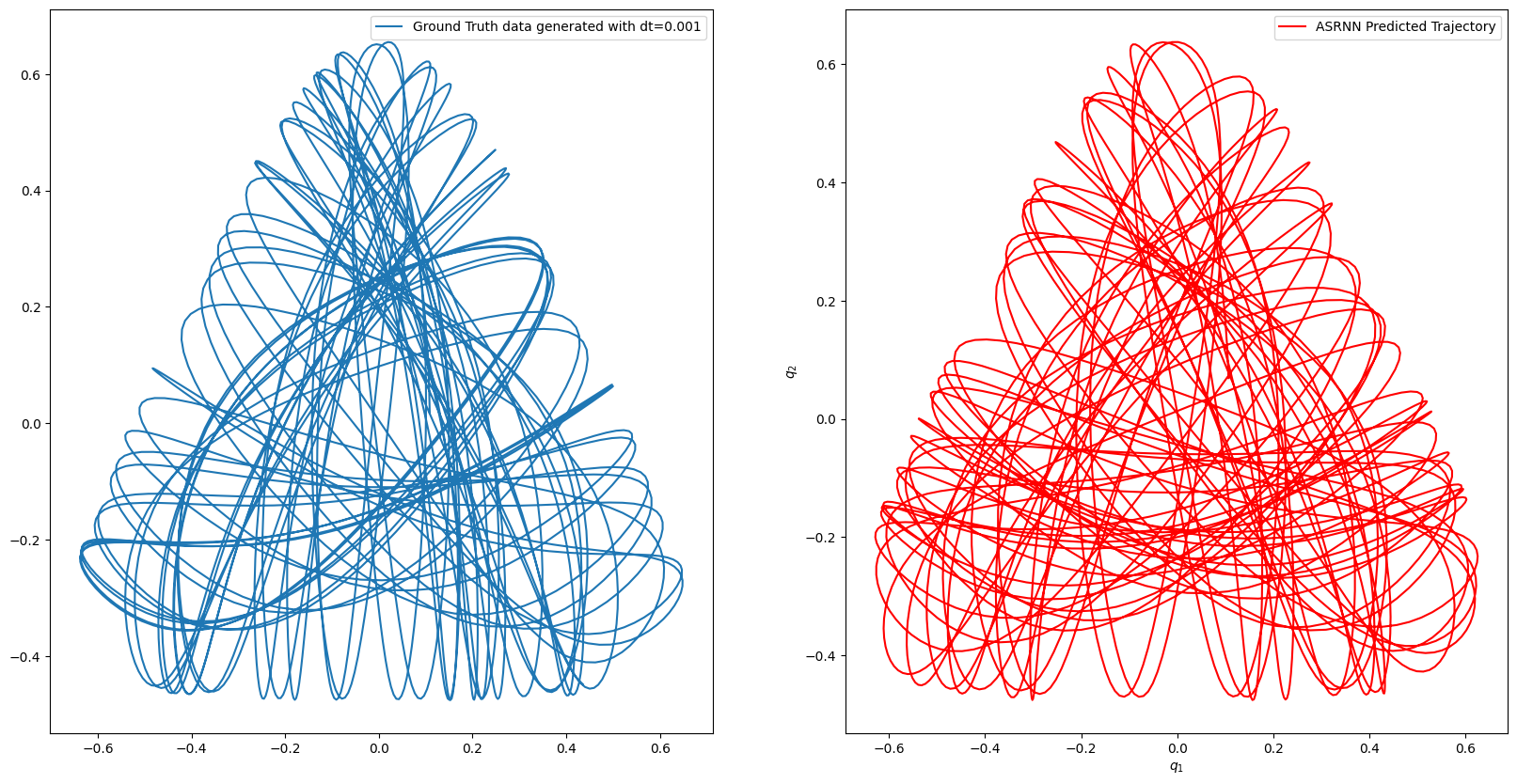}
    \caption{Predicted and true trajectories for $\begin{pmatrix}
        \alpha \\ \beta 
    \end{pmatrix} = \begin{pmatrix}
        1.0 \\ 0.8
    \end{pmatrix}$ with $E = \frac{1}{7}$}
    \label{fig:0.9_0.7highenergy}
\end{figure}

At this point it would be prudent to discuss the nature of the \textit{HNN conserved quantity}: this is the quantity w.r.t to which we are taking the gradients and we expect it to be conserved over time. We in fact that this quantity, referred to herein as $\mathcal{H_{\theta}}$, is conserved to a greater accuracy($\sim 10^{-4}$) than the energy. This suggests that $\mathcal{H_{\theta}}$, although similar to the Hamiltonian, is a distinct measure associated with the discretized version of the equations. Chen et al. \cite{Chen_Zhang_Arjovsky_Bottou_2020} suggest that the SRNN learns not the true underlying equations, but a modified version of the same associated with the Leapfrog integrator and time step size. In that sense the network is able to compensate for numerical discretization errors as well. They found, under certain conditions, that the SRNN outperforms even the simulation with true equations with the same step size, we however did not observe this with the Adaptable SRNN at this point. Greynadus et al. \cite{Greydanus_Dzamba_Yosinski_2019} also note that $\mathcal{H_{\theta}}$ diverges more slowly or not at all w.r.t to the total energy. However significant study on the nature and properties of $\mathcal{H_{\theta}}$ is required but challenging due to the black-box nature of neural networks \\

\subsection{Comparison of Results}

\begin{figure}[ht]
    \centering
    \subfloat{\includegraphics[width=0.9\textwidth]{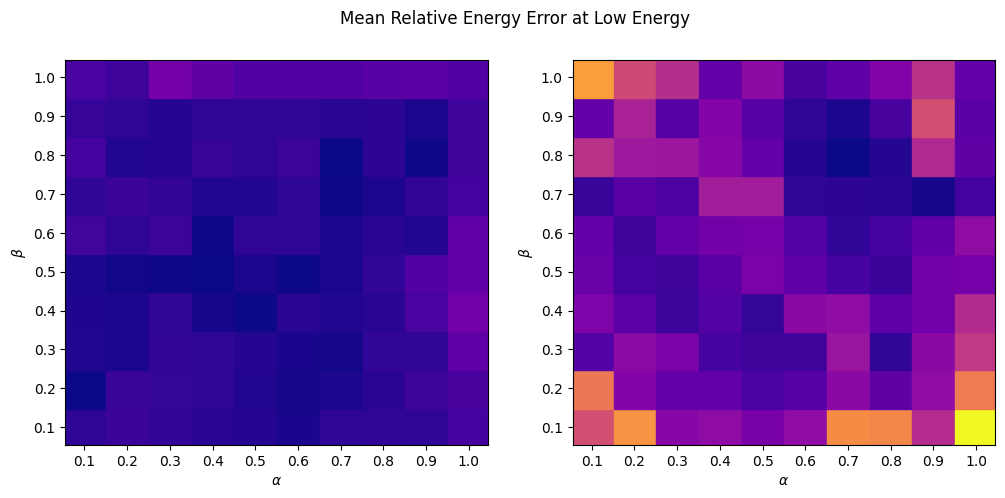}}
    \subfloat{\includegraphics[width=0.1\textwidth, height=6cm]{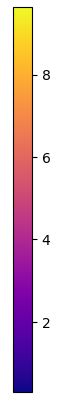}}
    \caption{$\langle \Delta E \rangle$  at low energies for left-ASRNN, right-AHNN}
    \label{fig:lowenergies}
\end{figure}

\begin{figure}[ht]
    \centering
    \subfloat{\includegraphics[width=0.9\textwidth]{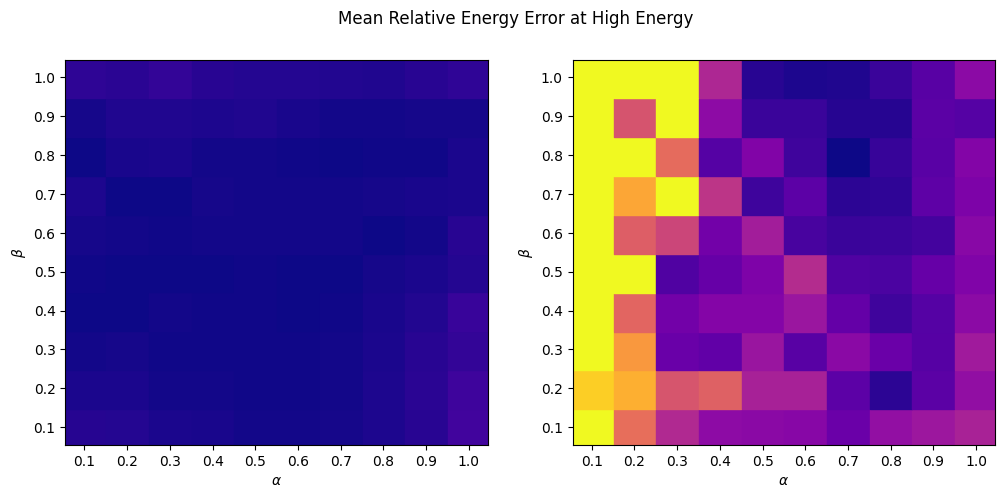}}
    \subfloat{\includegraphics[width=0.1\textwidth, height=6.5cm]{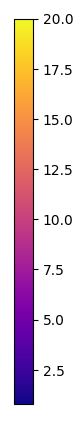}}
    \caption{$\langle \Delta E \rangle$  at high energies for left-ASRNN, right-AHNN}
    \label{fig:highenergies}
\end{figure}

\begin{figure}[ht]
    \centering
    \includegraphics[width=0.9\textwidth]{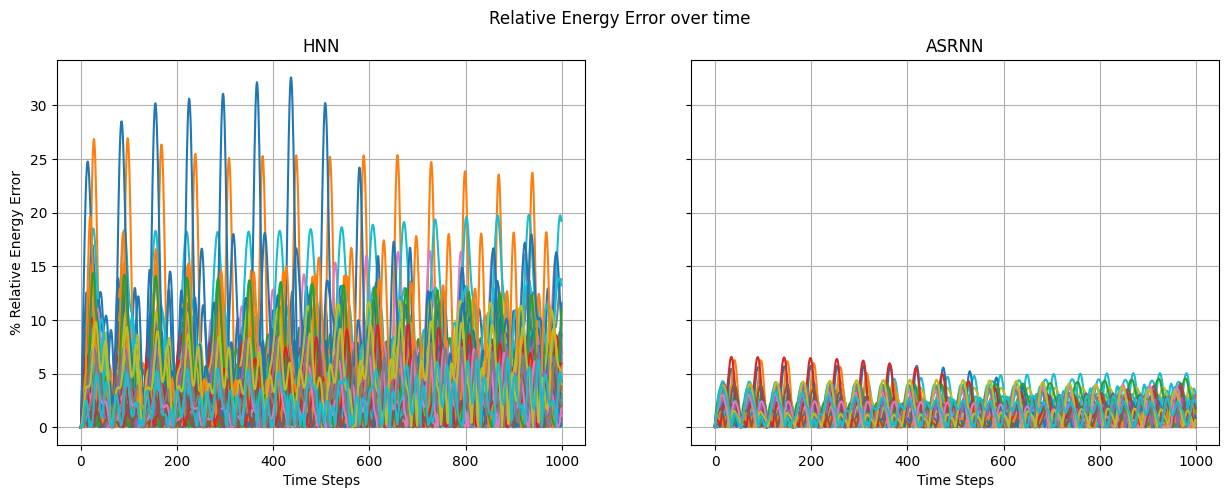}
    \caption{$\Delta E$ over time at low energy}
    \label{fig:Energytime}
\end{figure}

We now test the AHNNs and ASRNNs on the entire parameter space $\alpha, \beta \in (0, 1]$. It is found that in trajectories predicted upto $1000$ steps of separation $0.1$, the $\langle \Delta E \rangle$ is consistently lower for the ASRNNs vs the AHNNs staying below $2.5\%$ for the former. Figs. \ref{fig:lowenergies}, \ref{fig:highenergies}  show the error in energy over the entire parameter space. Especially at higher parameter values one finds significantly higher error for the latter. Further due to the enforcement of phase space volume conservation in the very structure of the ASRNN, error in energy doesn't grow with time as shown in Fig. \ref{fig:Energytime}, this however is not true for the AHNN. 

\begin{figure}[ht]
    \centering
    \subfloat{\includegraphics[width=0.9\textwidth]{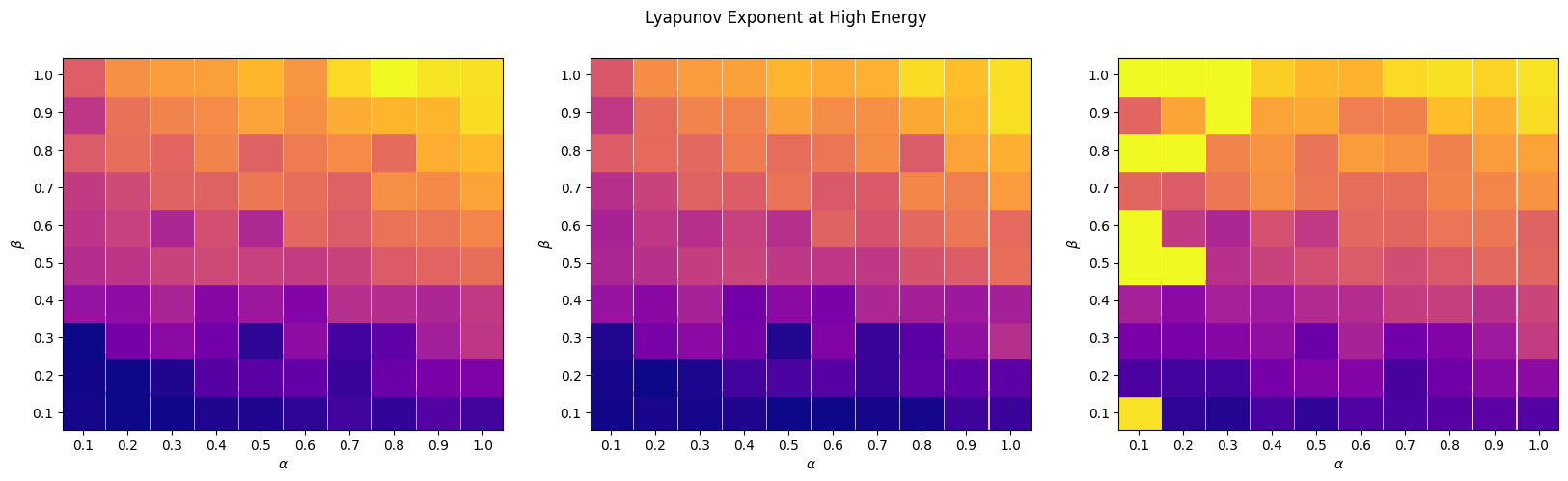}}
    \subfloat{\includegraphics[width=0.1\textwidth, height=4cm]{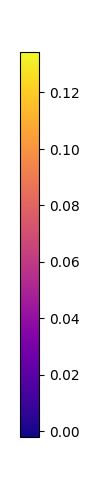}}
    \caption{Lyapunov Exponents over the Multiparameter Space in order True, ASRNN, AHNN from left-right}
    \label{fig:lyapmultiparam}
\end{figure}

One can see this in the context of the transition to chaos as well, we tested the Henon-Heiles for energies where we expect chaotic trajectories for certain parameter regimes, we then measure the Lyapunov exponents as we vary the parameters. For the entire $(\alpha, \beta)$ space, the Maximal Lyapunov exponent is calculated from an ensemble of predicted trajectories at high energies. Fig. \ref{fig:lyapmultiparam} shows clearly that the ASRNN's predictions match ground truth values very closely, while the AHNN diverges significantly beyond the training domain. \\

Hence ASRNNs are robust to large parameter spaces and initial conditions even for nonlinear and complex dynamics. Further they greatly outperform existing architectures to analyse Hamiltonian systems over a parameter set. They retain the statistical properties of the system such as Lyapunov exponents and are able to generalize from training data to learn dynamics beyond the training regime. In addition, their symplectic structure ensures conservation of energy to a great degree as well as obeying Liouville's theorem. In the following chapter we explore how these can be used to tackle the second proposed problem: discovering dynamics from partial information.

\chapter{Reconstructing Dynamics} \label{chapter4}
Having demonstrated the robustness of our proposed architecture, we now aim to solve our second proposed problem:

\begin{itemize}
    \item Consider a Hamiltonian dynamical system of phase space dimension $k$. We aim to reconstruct the entire phase space dynamics through data of the dynamics of a small number of degrees of freedom. We can consider the dynamics of the partially observable data to exist on a manifold of the phase space of dimension $d_p<k$ s.t. the dynamics of the entire system can be projected down to this manifold.
    \item Assuming training data is available, can we construct an architecture using the tools discussed which can not only reconstruct the entire dynamical system but also discover the parameters of the Hamiltonian while preserving Hamiltonian Dynamics.
\end{itemize}

\section{Takens' Embedding Theorem}

Takens' Embedding Theorem gives us the capability to construct a topologically invariant version of the original attractor of a dynamical system can from partial information, i.e.the \textit{Delay Embedding} of as less as one coordinate. The theorem is as follows:

\begin{enumerate}[I]
    \item First a point of definition: Consider a compact set $S$. The least $m \in \mathbb{N}$ s.t. $\exists$ a one-to-one mapping $F: S \to \mathbb{R}^m$ is defined as the \textbf{Embedding Dimension} of $S$. The mapping $F$ is referred to as the \textbf{Topological Embedding} of $S$.
    \item Consider as before a finite $\mathbf{k}$ \textbf{dimensional dynamical system} and let $\{S_t\}$ represent the time series of a single variable. The $\mathbf{m-dimensional}$ \textbf{Delay Vector} is defined as:
    \begin{equation}
        [S(t), S(t-\tau), S(t-2\tau), \hdots , S(t - (m-1)\tau)]
    \end{equation}
    For some fixed \textit{delay} $\tau$. In general we can construct an $m$ dimensional \textbf{delay plot} by graphing this vector of delay coordinates for each $t$ in the series. If $\exists$ a $m$ large enough s.t. $\exists$ a one-to-one mapping from set generated by the Delay Vector to $\mathbb{R}^m$ then that set is referred to as the \textbf{Delay Coordinate Embedding} of ${S_t}$.
    
    \item Takens' Theorem guarantees the existence of such an $m$ for a finite dimensional dynamical system and further provides an upper limit for it. Given our $k$ dimensional dynamical system with state space $\mathbb{R}^k$, lets say its trajectories are bounded to a $d$ dimensional submanifold where $d \leq k$. For Hamiltonian systems typically $d<k$ due to the energy conservation constraint which bounds the trajectory to a lower dimensional submanifold.  This submanifold is often referred to as the Attractor($A$) if it is invariant to the dynamical system. Lastly we define a measurement function $h : \mathbb{R}^k \to \mathbb{R}$ which defines the process of measuring(or projecting to) $S$. Given the above, Takens' theorem states that:

    \textit{Assume that $A$ is a d-dimensional submanifold of $\mathbb{R}^k$ which is invariant to the dynamical system $\mathbf{X}$. If $m>2d$ and $\mathbf{F} : \mathbb{R}^k \to \mathbb{R}^m$ is a delay coordinate function wih a generic measurement function $h$ and generic time delay $\tau$, then $F$ is one-to-one on $A$.} 

    Essentially Takens' Theorem tells that if the attractor dimension is $d$ then the embedding dimension for the delay coordinates is at most $2d+1$.

\end{enumerate}

Uribarri and Mindlin \cite{Uribarri2022Jan} showed that appropriate LSTM architectures appear to learn how to implement Takens' Embedding Theorem, i.e. they learn to generate an embedding of the data in their own hidden/cell state which is topologically invariant to the original attractor. They demonstrate this using an encoder-decoder architecture and using linking numbers to demonstrate topological invariance. Recently Young and Graham\cite{PhysRevE.107.034215} studied the problem of learning delay coordinate dynamics using deep neural networks. They use these to construct a map from the partially observable data to the delay embedding following which they train a network to learn the mapping from the embedding to the true attractor. \\

We aim to use the former idea to generate a topologically invariant embedding of a Hamiltonian system with an LSTM which we couple with a Hamiltonian Neural Network architecture in order to enforce Hamiltonian Dynamics. We have shown that the ASRNN is far more robust than the AHNN as a Hamiltonian Dynamics informed architecture in both single parameter and multiparameter conditions, hence we use this to couple with the LSTM.

\section{LSTM Encoding Architecture}

\begin{figure}[ht]
    \centering
    \includegraphics[width=0.8\textwidth]{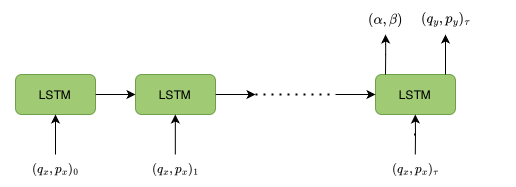}
    \caption{Architecture of Encoding LSTM}
    \label{fig:lstmencoder}
\end{figure}

We now construct a similar LSTM architecture whose aim is to embed the information of the system obtained from the time series of a few(or even one) variables in the LSTM hidden state s.t. it is able to predict the parameters of the Hamiltonian. Lets say we provide the network with only the time series of $(q_x, p_x)$ and aim to predict $(q_y, p_y)$ at the last time step along with $(\alpha, \beta)$, i.e. we are aiming to not only reconstruct the topologically invariant attractor but also find the mapping from the same to the canonical variables. The proposed architecture is shown in Fig. \ref{fig:lstmencoder}. The ability of the LSTM to learn what information to keep or discard, helps us have considerable freedom on the length of the input time series: we need it only to be sufficiently long to allow for a complete embedding. The maximum embedding dimension for the Henon Heiles is $m=7$, this is because the constraint on energy restricts the dynamics of the system to a $3$ dimensional manifold. Normally one of the great challenges of using Takens' embedding theorem in practise is determining the optimum embedding dimension, however we note that an LSTM has the ability to learn this dimension due to its ability to control how much information is stored in its hidden state and how much is taken from previous time steps. We hence fix the size of the hidden state of the LSTM to $9$: a slightly greater value than the maximum embedding dimension $7$ to allow it freedom to embed the dynamics in a higher or lower dimensional space which may make it easier to determine the mapping to canonical variables. Hence a time series of length $\tau$ is used to generate the embedding which is then mapped to the output and the parameter is recovered. The loss function for this network is hence:

\begin{equation}
    \mathcal{L}_{(q_x, p_x)}(\theta) = \norm{\alpha - \hat{\alpha}}_2 + \norm{\beta - \hat{\beta}}_2 + \norm{q_y - \hat{q}_y}_2 + \norm{p_y - \hat{p}_y}_2
\end{equation}

\begin{figure}
    \centering
    \includegraphics[width=\textwidth]{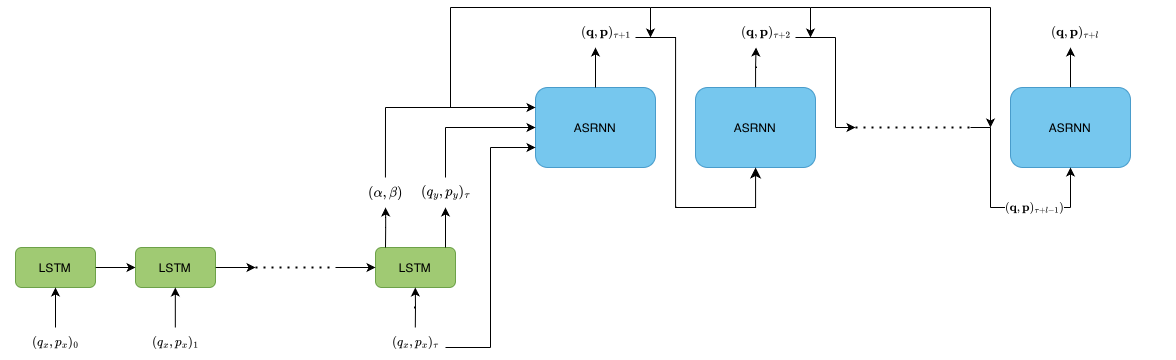}
    \caption{Architecture of a Coupled LSTM-ASRNN}
    \label{fig:LSTMASRNN}
\end{figure}

We first investigate the performance under single parameter conditions, the networks are constructed and trained as follows:

\begin{itemize}
    \item Time series for $(\mathbf{q}, \mathbf{p})$ are generated exactly as in Sec. \ref{single parameter SRNN} for the same parameter and energy values.
    \item After trying several different encoding sequence lengths, a length of $30$ steps separated with time step $0.1$ was fixed for all batches. Hence for each $(q_y, p_y)$ point, the previous 30 steps of $(q_x, p_x)$ were taken as the input encoding sequence.
    \item The internal structure of the ASRNN is kept same as before while the LSTM is constructed with a single layer followed by a linear transformation to output.
    \item The two networks are trained independently of each other with the same data then coupled in the form shown in Fig. \ref{fig:LSTMASRNN} to allow prediction with partial information.
\end{itemize}

\begin{figure}
    \centering
    \subfloat[$\alpha = 0.3$, $\mu_{\alpha} = 0.33$, $\sigma =0.12$]{\includegraphics[width=0.33\textwidth]{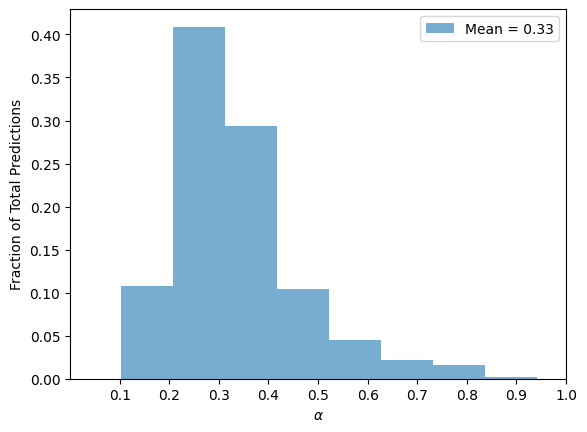}}
    \subfloat[$\alpha = 0.5$, $\mu_{\alpha} = 0.51$, $\sigma =0.13$]{\includegraphics[width=0.33\textwidth]{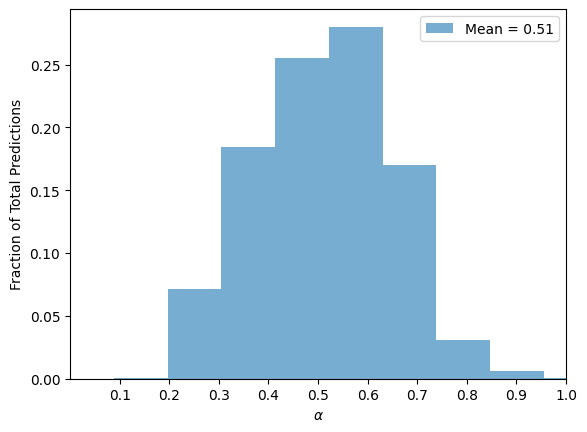}}
    \subfloat[$\alpha = 0.7$, $\mu_{\alpha} = 0.68$, $\sigma =0.15$]{\includegraphics[width=0.33\textwidth]{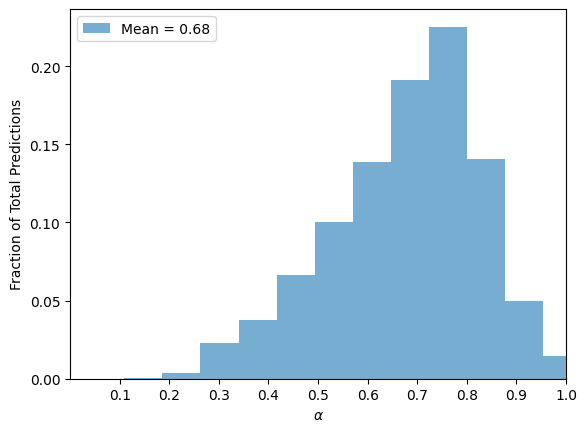}}    
    \caption{Distribution of predicted Values of $\alpha$}
    \label{fig:alphadistributions}
\end{figure}

The LSTM predicts the parameter to a reasonably high accuracy: lets say we feed several trajectories corresponding to various initial conditions with the same parameter $\alpha$ and then attempt to quantify the accuracy of the networks prediction with the mean and standard deviation of independent predictions of $\alpha$. Fig. \ref{fig:alphadistributions} shows some histograms displaying the distributions of predicted values, we find that the standard deviation($\sigma$) of $\alpha$ for values in the entire range was approximately $0.13$ with a maximum of $0.16$. It should be noted that the LSTM is only trained for 4 distinct parameter values yet is able to generalize to the entire interval. \\

We now layer this trained LSTM with the ASRNN trained earlier to predict solely from the $(q_x, p_x)$ time series.

\section{Results}

The LSTM-ASRNN combination performs well but requires considerable study to optimize, however the results presented here do demonstrate the promise of this architecture. For brevity we report the results for the high energy and low energy for a few parameter values, the behaviour is largely the same over the entire parameter space when studied statistically. As we saw in Fig. \ref{fig:alphadistributions}, often times the LSTM predicts wildly different values of the parameter, this can lead to very different predictions by the ASRNN. However if multiple similar trajectories(same parameter) are fed to the LSTM, the mean of predicted values is typically very close to the true parameter. Hence this method has to be followed when predicting with the layered architecture; in practice this can be thought of as dividing the available data for a trajectory(which of course has the same parameter) into smaller slices which are fed as independent inputs to the LSTM. The mean of the $\alpha$ predictions by the LSTM are then fed to the ASRNN for predicting the rest of the dynamics. \\

\begin{figure}
    \centering
    \includegraphics[width=\textwidth]{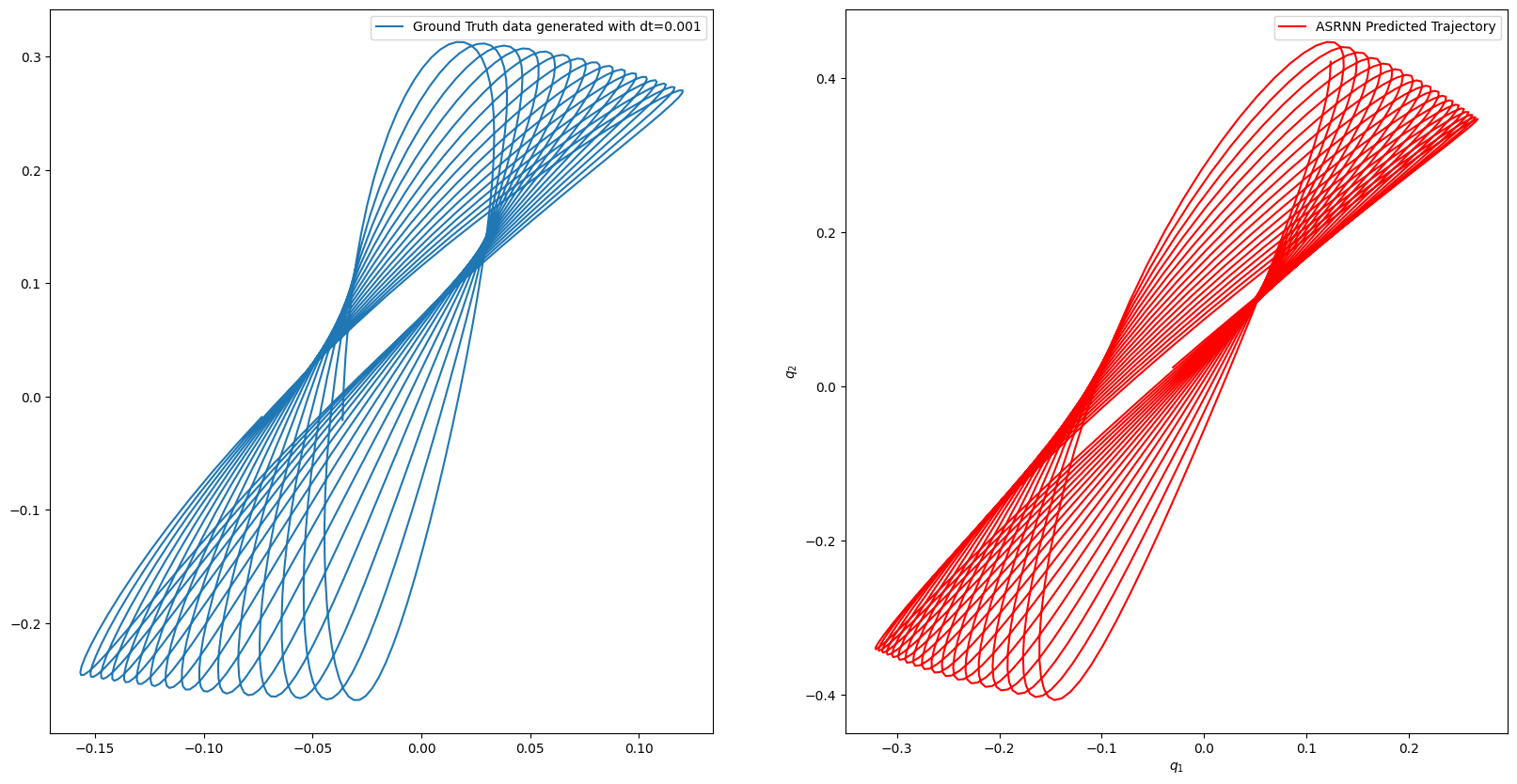}
    \caption{Comparison of trajectories predicted at low energies with true $\alpha = 0.3$}
    \label{fig:partialinfo1}
\end{figure}

\begin{figure}
    \centering
    \includegraphics[width=\textwidth]{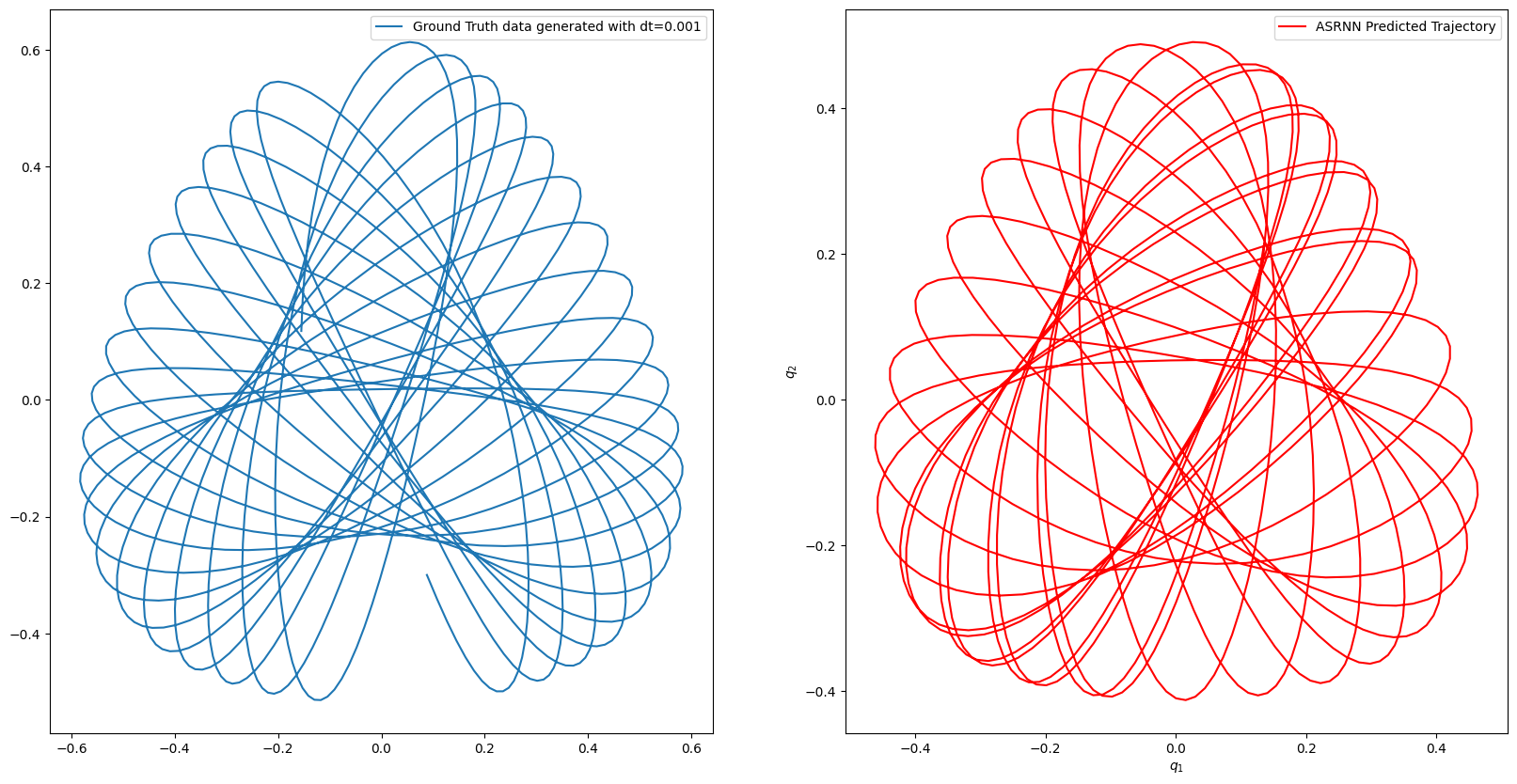}
    \caption{Comparison of predicted trajectories at high energies with true $\alpha = 0.5$}
    \label{fig:partialinfo2}
\end{figure}

\begin{figure}
    \centering
    \includegraphics[width=\textwidth]{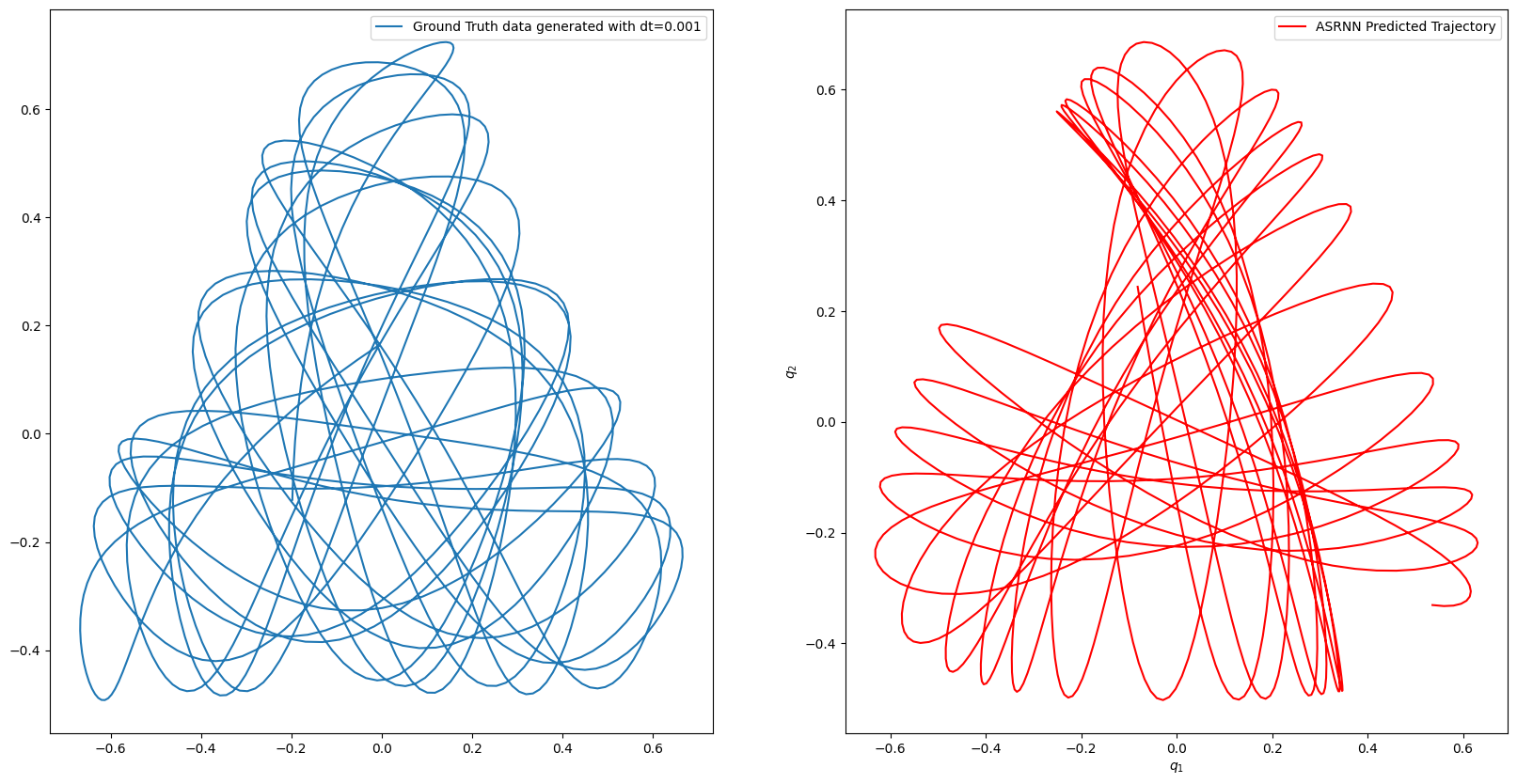}
    \caption{Comparison of predicted trajectories at high energies with true $\alpha = 1.0$}
    \label{fig:partialinfo3}
\end{figure}

In general we find that this method performs well at both high and low energies over the entire parameter space. Fig. \ref{fig:partialinfo1} shows an example trajectory at low energy, the mean predicted $\alpha$ value for this case was $0.78$. Similarly Figs. \ref{fig:partialinfo2} and \ref{fig:partialinfo3} show this for high energies at two different values of $\alpha$. It should be noted that for two parameters, the LSTM is unable to predict them to a good enough accuracy. Hence, even though the ASRNN is robust to multiparameter potentials, the proposed architecture is not efficient to study them.\\

Presented here is only a preliminary analysis of this method which suggests that an architecture similar to the one proposed can successfully tackle the problem of learning the dynamical system from partial information. Although this work does not cover these aspects, further study on this can include training the two networks as a single unit rather than separately as done here and developing methods to effectively embed the series in the LSTM hidden space. The latter requires using existing methods to determine the optimal embedding dimension and delay spacing. Further one can make use of Deep neural networks instead of linear transformations to potentialy better learn the map from the delay embedding to the true attractor.

\chapter{Conclusion} \label{conclusion}
This dissertation studied the capability of Neural Networks to model and predict Hamiltonian Dynamical systems while taking into account the symplectic structure of Phase space and conservation of a first integral of motion. Specifically we proposed an extension of existing Hamiltonian Neural Network architectures which we refer to as Adaptable Symplectic Recurrent Neural Networks(ASRNNs). The performance of this architecture is investigated using the Henon-Heiles potential as a toy model of a nonlinear and potentially chaotic Hamiltonian and the results are compared to those obtained from existing structures. These ASRNNs are designed learn the dynamics of Hamiltonian systems from data of canonical coordinates and hence to predict the trajectory of a Hamiltonian system at a range of given parameter values and initial conditions. They encode the leapfrog algorithm in their very structure and hence preserve the volume of phase space over time. It is demonstrated that ASRNNs are capable of learning and effectively predicting trajectories in the entire parameter space given training information of only a few distinct points. \\

Two primary measures are used for the analysis: relative error in energy from the true trajectory and the Lyapunov exponent. It is found that ASRNNs conserve energy to a far greater degree than existing architectures. Further it is seen that they are able to learn the transition to chaos given data of mostly quasiperiodic orbits. Further research with ASRNNs includes potentially applying them to systems with other symmetries, such as spherical, and attempting to learn dynamics while respecting the other integrals of motion as well. Additionally, this work does not study ASRNNs while taking into account noise, hence an implementation of the architecture alongside Initial state optimization, presented by Chen et al. \cite{Chen_Zhang_Arjovsky_Bottou_2020} as a method to handle noise in SRNNs, would help better their robustness to noise. It is stressed that \textbf{ASRNNs greatly outperform existing architectures in the context of learning Hamiltonian Dynamics over a range of parameter values}.\\

The second problem this work tackles is that of learning a Hamiltonian system given only partial information of its dynamics. Motivated by the work done on the Rossler System by Uribarri and Mindlin \cite{Uribarri2022Jan}, an LSTM is used as a method to create a delay embedding using data of a few canonical coordinates. A Dense neural network is then used to map the delay embedding to the true phase space while also extracting the value of the parameter for a single parameter potential. This LSTM architecture is then layered with an ASRNN to learn the dynamics of the system and make future predictions while obeying Hamilton's equations. Only a preliminary analysis of this method, which shows its potential, is presented and further research is proposed. \\

Overall this work shows the potential to significantly enhance the performance of Neural Networks when the laws of physics are combined with them. It has been demonstrated that using the capability of neural networks to learn high dimensional dynamics along with traditional integrators such as Leapfrog designed to obey the laws of physics, can help the networks converge to solutions which take these laws into account. With the growing computing capabilities available today, the possibility of developing such architectures has emerged  as a compelling method to discover new physical laws using empirical data and existing knowledge of physics.

\appendix
\chapter{Appendix}
\section{The Backpropogation Algorithm} \label{backprop}

As discussed in Sec. \ref{feedforward}, calculating the gradients of the loss function with respect to the parameters using finite differences is a computationally impossible task due to significant number of parameters. However one can take advantage of the chain rule of multivariable calculus to simplify this calculation significantly. We need only consider a single input and find the gradient w.r.t to it, hence in the following we redefine $N$ to represent the number of layers, and the sum to define mean the difference between the ground truth output and that obtained with input $x_i$. Thus the sum is now over the outputs of the network.  \\

The loss function is given by
\begin{equation}
    \mathcal{L}_x(\theta) = \frac{1}{N} \sum_ i\left( x_i^N - F(x_i)\right)^2
\end{equation}
Where $F_\theta(x_i)$ has been replaced by $x_i^n$, since the former is simply the neuron activations for the final($Nth$) layer. Consider the partial derivative of the loss function w.r.t to some arbitrary weight or bias $w^*$: using the chain rule this can be written as:
\begin{equation}
    \frac{\partial \mathcal{L}_x(\theta)}{\partial w^*} = \sum_{i} \left( x_i^N - F(x_i)\right) \frac{\partial x_i^N}{\partial w^*}
\end{equation}
Note that we have suppressed the $1/N$ factor for simplicity since its just a constant. Since $x_i^N = f(z_j^N)$, with another application of the chain rule this can be written as:
\begin{equation}
    \frac{\partial \mathcal{L}_x(\theta)}{\partial w^*} = \sum_{i} \left( x_i^N - F(x_i)\right) f'(z_i^N) \frac{\partial z_i^N}{\partial w^*}
\end{equation}
But $z_j^{N} = \sum_k w_{jk}^{N, N-1}x_k^{N-1} + b_j^{N} $, hence
\begin{equation}
    \frac{\partial z_i^N}{\partial w^*} = \sum_j \frac{\partial z_i^N}{\partial y_j^{N-1}} \frac{\partial y_j^{N-1}}{\partial w^*}  = \sum_j \frac{\partial z_i^N}{\partial y_j^{N-1}}f'(z_j^{N-1}) \frac{\partial z_j^{N-1}}{\partial w^*}
\end{equation}
Clearly $\frac{\partial z_k^{N-1}}{\partial w^*}$ can be expanded into a sum of partial derivatives of the activations of the previous layer like how we did for $\frac{\partial z_i^N}{\partial w^*}$. In general for the $nth$ layer we can write:
\begin{equation}
    \frac{\partial z_i^n}{\partial w^*} =  \sum_j \frac{\partial z_i^n}{\partial y_j^{n-1}}f'(z_j^{n-1}) \frac{\partial z_j^{n-1}}{\partial w^*}
\end{equation}
But $\frac{\partial z_i^n}{\partial y_j^{n-1}} = w_{ij}^{n, n-1}$, hence 
\begin{equation}
    \frac{\partial z_i^n}{\partial w^*} =  \sum_j  w_{ij}^{n, n-1}f'(z_j^{n-1}) \frac{\partial z_j^{n-1}}{\partial w^*}
\end{equation}
Hence every $\frac{\partial z_i^n}{\partial w^*}$ can be calculated as a sum of the partial derivatives of $z$ of the previous layer. Hence starting from the top layer, we can go backward through the layers until we reach the weight/bias we wanted to calculate the partial derivative w.r.t, in which case the backward pass will terminate. Let $w^* = w_{jk}^{\Tilde{n}, \Tilde{n}-1}$ for some $j, k, \Tilde{n}$, then:
\begin{equation}
    \frac{\partial z_j^{\Tilde{n}}}{\partial w^*} = x_k^{\Tilde{n}-1}
\end{equation}
Or if $w^*$ was a bias, i.e. $w^* = b_{j}^{\Tilde{n}} \Rightarrow \frac{\partial z_j^{\Tilde{n}}}{\partial w^*} = 1$. Hence $\Tilde{n}$ is the layer at which the backward pass would terminate. \\

Thus the algorithm boils down to sucessive matrix multiplications through the net until we reach the layer corresponding to the parameter being differentiated against, making it significantly faster than the finite differences alternative. The algorithm can be formulated explicitly in terms of matrices as follows: \\

Define the matrix $M_{ij}^{n, n-1} = w_{ij}^{n, n-1} f'(z_j^{n-1}$, then $\frac{\partial z_i^n}{\partial w^*} = \sum_j M_{ij} \frac{\partial z_i^{n-1}}{\partial w^*}$. Hence:
\begin{equation}
    \frac{\partial z_i^N}{\partial w^*} = \sum_{j, k, \hdots, u, v} M_{ij}^{N, N-1} M_{jk}^{N-1, N-2} \hdots M_{uv}^{\Tilde{n+1, \Tilde{n}}} \frac{\partial z_v^{\Tilde{n}}}{\partial w^*}
\end{equation}
Which gives us the gradient w.r.t to the loss function:
\begin{equation}
    \frac{\partial \mathcal{L}_x(\theta)}{\partial w^*} = \frac{1}{N}\sum_{i} \left( x_i^N - F(x_i)\right) f'(z_i^N) \frac{\partial z_i^N}{\partial w^*}
\end{equation}
To calculate the complete gradient we have to do this for every weight and bias, however the primary advantage of this algorithm becomes that the different partial derivatives differ only in the partial derivative of the last term, hence one backward pass is all that is necessary to find the entire gradient. \\

With the gradient calculated, the parameters of the neural network are updated by Stochastic gradient descent:

\begin{equation}
    \theta \to \theta - \eta \nabla_{\theta} \Tilde{L}_x(\theta)
\end{equation}
Where $\Tilde{L}$ represents a randomly sampled subset of batches, hence the name `stochastic'. $\eta$ defines a `learning rate' which scales how large a step is taken in the calculated direction, it may be constant or variable. The idea is that the noise of randomly selecting a batch cancels out over sufficiently many steps if $\eta$ is small enough.  This process of gradient descent is performed iteratively to `step', or rather random walk, towards the minima.

\section{Algorithm to Calculate Lyapunov Exponents} \label{lyapunovalgorithm}

Lyapunov exponents($\lambda$) are a measure of the stretching/contraction of nearby trajectories, i.e. they measure the rate at which trajectories corresponding to close-by initial conditions diverge over time. For chaotic systems $\lambda>0$ while for integrable systems $\lambda \leq 0$. Typically there exists a Lyapunov exponent for each independent degree of freedom of a dynamical system, in most case we define the system as chaotic or integrable on basis of the \textit{Maximal Lyapunov exponent}. A naive method to calculate these would be to simply start with two close initial conditions and measure the difference between their trajectories over time, however especially for multidimensional systems, it is better to take a slightly more complex route. Lets first define the Lyapunov exponent: \\

\textit{Let $f$ be a smooth map on $\mathbb{R}$. The Lyapunov exponent $\lambda(x_1)$ of the orbit $\{x_1, x_2, \hdots\}$ is defined as:}
\begin{equation}
    \lambda(x_1) = \lim_{n\to\infty} \frac{1}{n} \sum_i \ln|f'(x_i)|
\end{equation}
\textit{if this limit exists.}
It is straightforward to extend this definition to flows: \textit{The Lyapunov exponent for a flow is defined as the exponent for its \textbf{Time-1} map given by map $F(\mathbf{v}(t)) = F(\mathbf{v}(t+1))$, i.e. the image of the point $\mathbf{v}$ is the position of the solution at time $1$ of the initial value problem with $\mathbf{v_0} = \mathbf{v}$}. \\

Consider a set of coupled differential equations $\Dot{\mathbf{v}} = f(\mathbf{v})$. Imagine now an orthonormal basis $\{\mathbf{w}_1^0, \mathbf{w}_2^0, \hdots \mathbf{w}_d^0\}$ where $d$ is the dimension of the system. A step forward in time will lead to these vectors expanding or contracting by an amount determined by the Jacobian($\mathbf{J}$) of the flow, hence let the vectors $\mathbf{z}_1, \hdots, \mathbf{z}_d$ be defined as:
\begin{equation}
    \mathbf{z}_1 = \mathbf{J}(\mathbf{v}_0) \mathbf{w}_1^0, \hdots, \mathbf{z}_d = \mathbf{J}(\mathbf{v}_0) \mathbf{w}_d^0
\end{equation}
We are interested to know how much these vectors have expanded or contracted, or equivalently how the volume of the parallelopiped formed by the vectors has changed. To get an estimate of the change in volume over time, it is necessary to do this successively over many steps. However, these resulting vectors $\mathbf{z}_i$ are not necessarily orthogonal. We hence use the Gram-Schmidt Orthogonalization procedure(which preserves the determinant) to get another set of orthogonal vectors say $\{\mathbf{y}_1, \mathbf{y}_2, \hdots \mathbf{y}_d\}$. To prevent the vectors from growing arbitrarily large or small, this new set is renormalized before applying the procedure again, i.e. define $\mathbf{w}_i^1 = \frac{\mathbf{y_i}}{\norm{\mathbf{y}_i}}$. Then we recalculate how much this new basis expands by due to the flow:
\begin{equation}
    \mathbf{z}_1 = \mathbf{J}(\mathbf{v}_1) \mathbf{w}_1^1, \hdots, \mathbf{z}_d = \mathbf{J}(\mathbf{v}_1) \mathbf{w}_d^1
\end{equation}
Which produces another set of $\mathbf{y}$ vectors and correspondingly $\mathbf{w}$ basis. It is clear that $\norm{y_i}$ measures the one step growth in the $ith$ direction, hence the $ith$ Lyapunov exponent can be calculated as a mean of $\ln \norm{\mathbf{y}_i}$ over time:
\begin{equation}
    \lambda_i = \frac{1}{N}\sum_n \ln \norm{\mathbf{y}_i^n} 
\end{equation}
Where $N$ is the number of steps.\\

Hence to find the Lyapunov exponent we need the Jacobian of the system, in principle this can be calculated for our Hamiltonian Neural networks by calculating second order partial derivatives w.r.t to $(\mathbf{q}, \mathbf{p})$ by automatic differentiation. However this proves to be computationally expensive, one can hence approximate $\mathbf{J}$ more efficiently using finite differences. In practice, the vectors mentioned above are calculated in terms of matrix multiplications: \\

Let $\mathbf{W}$ be a matrix storing the vectors $\mathbf{w}$, hence $\mathbf{W} = I$ initially. Hence according to our algorithm:
\begin{equation}
    \mathbf{Z}(t) = \mathbf{J(t_0) \cdot \mathbf{W}(0)}
\end{equation}
We can easily calculate the Gram-Schmidt orthogonalization using the \textit{QR Decomposition} of $\mathbf{Z}$, this gives us:
\begin{align}
    \norm{\mathbf{Y}} = Diag(R) \\
    \mathbf{W}(t+1) = Q
\end{align}
Where $Diag(A)$ represents the main diagonal of the matrix A. Thus we can sucessively calculate $\norm{\mathbf{Y}}$ and estimate the Lyapunov exponent as defined before. Finally we define the maximal Lyapunov exponent as
\begin{equation}
    \lambda_M = \max_i(\lambda_i)
\end{equation}
\singlespacing
\printbibliography

\end{document}